\documentclass{article}

\usepackage{iclr2027_conference,times}

\usepackage{amsmath,amsfonts,bm}

\def\eqref#1{equation~\ref{#1}}

\def\1{\bm{1}}

\DeclareMathAlphabet{\mathsfit}{\encodingdefault}{\sfdefault}{m}{sl}
\SetMathAlphabet{\mathsfit}{bold}{\encodingdefault}{\sfdefault}{bx}{n}

\usepackage{amsmath,amssymb,amsthm}
\usepackage{booktabs}
\usepackage{graphicx}
\usepackage{subcaption}
\usepackage{flafter}
\usepackage{xcolor}
\usepackage{colortbl}
\usepackage{hyperref}
\usepackage{url}

\renewcommand{\iclrruler}[1]{}

\newtheorem{proposition}{Proposition}
\newcommand{\method}{\textsc{Prism}}

\newcommand{\mssmdemos}{500}
\newcommand{\daggerdemos}{100}
\definecolor{simtableheader}{HTML}{E8EEF7}
\definecolor{simtableprism}{HTML}{EAF5EF}
\title{\raisebox{-0.24em}{\includegraphics[height=1.24em]{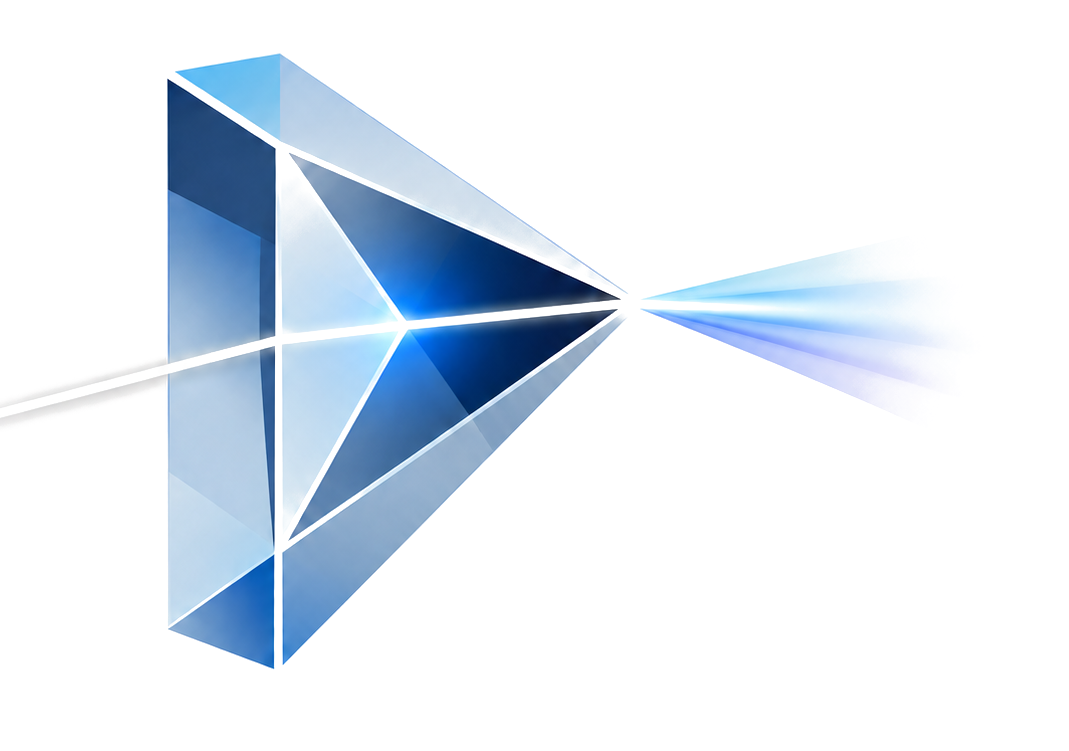}}\hspace{0.03em}%
PRISM: Precision Forcing for Generalist Vision-Language-Action Policies}
\author{%
Chenghao Yue$^{2}$, Peiyan Li$^{2}$, Hao Wang$^{3}$, Lei Lv$^{4}$, Jiahao Chen$^{5}$,\\
\textbf{Huaping Liu}$^{2,\dagger}$, \textbf{Yu Luo}$^{1,\dagger}$, \textbf{Yuanlong Zhang}$^{2,\dagger}$\\[0.35em]
\normalfont\small
$^{1}$Seen-E Robotics \quad $^{2}$Tsinghua University \quad
$^{3}$Dalian University of Technology\\
$^{4}$Tongji University \quad $^{5}$Peking University\\[0.35em]
\normalfont\small
\href{https://seen-e.github.io/PRISM/}{\textcolor{magenta}{\texttt{https://seen-e.github.io/PRISM/}}}
}

\begin{document}
\raggedbottom
\maketitle
\begingroup
\renewcommand{\thefootnote}{\fnsymbol{footnote}}
\footnotetext[2]{Corresponding authors}
\endgroup

\begin{abstract}
Coupling a pretrained vision-language model (VLM) with a diffusion or flow-matching action expert has become the mainstream vision-language-action (VLA) paradigm for embodied policies, yet they remain unreliable once a task demands precise manipulation, e.g., submillimeter cable insertion. We attribute this weakness to a training objective that averages flow-matching errors across contexts and flow times, allocating little supervision to the few decisive moments determining task success. Along the generative path, inference integrates a locally supervised velocity field, so equally penalized errors at different flow times can affect the executed action unequally. Consequently, precision-task performance deteriorates as physical tolerance tightens by orders of magnitude. We call this supervision imbalance across contexts and flow times \emph{Flow-Time Starvation}. The remedy is to jointly align supervision allocation and representation learning with context-dependent precision demand, enabling the policy to recognize when greater precision is needed. We instantiate this precision-forcing strategy as Precision-aware Reweighting and Internalization for Skilled Manipulation (\method{}), which scores precision demand offline, strengthens flow supervision where errors are consequential and action signals are visible, and internalizes that demand in the VLM condition representation. Simulation and real-robot experiments show that \method{} preserves or improves general-task performance while raising mean end-to-end success from $38.3\%$ to $86.7\%$ across drive connector insertion, bimanual connector mating, and screw tightening, substantially extending the precision boundary of existing embodied policies without new sensors or observation modalities.
\end{abstract}

\section{Introduction}
\label{sec:introduction}

\begin{figure*}[t]
  \centering
  \includegraphics[width=\linewidth]{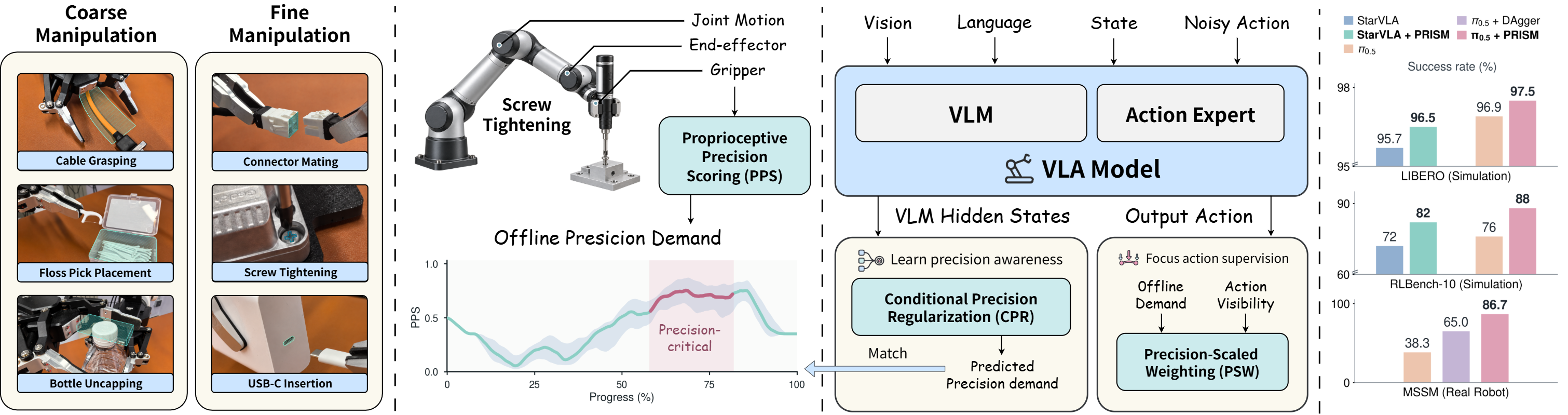}
  \caption{\textbf{Overview of \method{}.} PPS estimates precision demand from robot trajectories. CPR learns precision-aware VLM features, while PSW focuses action supervision on precision-demanding contexts with visible action evidence, improving fine manipulation without changing policy inference.}
  \label{fig:teaser}
\end{figure*}

Vision-language-action (VLA) policies have emerged as a central approach to general-purpose embodied control, building on a sequence of advances in robot policy learning. Action chunking and iterative generative modeling established temporally coherent continuous control~\citep{act,diffusionpolicy,octo}, and large-scale robot data paired with vision-language representations extended this capability to open-ended instruction following~\citep{rt1,rt2,openvla}. Coupling a pretrained vision-language model (VLM) with a diffusion or flow-matching action expert has since become a mainstream VLA paradigm~\citep{pi0,pi05,smolvla,grootn1,qwenrobotmanip}. Despite broad competence acquired through VLM pretraining and robot post-training, these policies still struggle with precise operations such as inserting a connector into a socket or fastening a screw.

We investigate this precision gap by examining whether training penalties reflect the physical consequences of action errors. Routine motions often tolerate centimeter-scale positional deviations, whereas precise interactions can require millimeter or sub-millimeter accuracy. Yet comparable errors in normalized action space receive similar penalties, even though the same physical deviation can be harmless during transport and decisive during alignment. Brief precision-critical interactions can therefore determine task success while receiving a smaller share of trajectory-averaged supervision than the longer routine motions around them~\citep{keyframeil,juicer}. This discrepancy also extends along the generative path, where inference integrates a velocity field learned from local supervision. Consequently, residuals with equal training penalties can have unequal effects on the executed action because their propagation depends on flow time. We formalize this mismatch between supervision and physical consequence across task contexts and flow times as \emph{Flow-Time Starvation} (see Section~\ref{sec:problem_formulation} for details).

These findings suggest supervising not only the demonstrated action, but also the precision that its execution requires. The resulting training signal should both concentrate learning on consequential errors and make precision demand recognizable from the policy's condition. We implement this \emph{precision-forcing} principle with Precision-aware Reweighting and Internalization for Skilled Manipulation (\textbf{\method{}}). Proprioceptive Precision Scoring (PPS) estimates continuous frame-level demand offline from robot trajectories. Precision-Scaled Weighting (PSW) couples this demand with flow-time emphasis and a signal-to-noise ratio (SNR) gate to concentrate additional supervision where action evidence is visible. Conditional Precision Regularization (CPR) makes the same demand predictable from the VLM condition representation. \method{} changes only training, leaving the model architecture and inference procedure unchanged.

Evaluation of \method{} spans standard simulation benchmarks and three extended real-robot tasks with precision-critical interactions. We refer to this setting as Multi-Stage Submillimeter-Tolerance Manipulation (MSSM). The experiments test whether precision forcing extends the precision boundary without sacrificing general breadth. \textbf{\method{}} preserves or improves general-task performance and raises mean end-to-end success on MSSM from $38.3\%$ to $86.7\%$.

Our contributions are summarized below.
\begin{itemize}
 \item \textbf{A formal diagnosis of precision undertraining.} We identify and formalize \emph{Flow-Time Starvation} through an error-propagation bound that accounts for physical tolerance and flow time. Thus, equally penalized velocity errors can affect success differently.
 \item \textbf{A precision-forcing training strategy.} We introduce \textbf{\method{}}, where data-driven PPS estimates frame-level precision demand before training, PSW combines this demand with flow-time emphasis and SNR-based visibility, and CPR encourages the VLM condition representation to encode precision demand.
 \item \textbf{Evaluation across breadth and precision.} We evaluate \method{} on standard simulation benchmarks and three MSSM real-robot tasks, demonstrating preserved or improved general-task performance and higher success on precision-critical interactions.
\end{itemize}

\section{Problem Formulation}
\label{sec:problem_formulation}

Let $c=(o,l,s)$ denote the current visual observation, language instruction, and proprioceptive state, and let the policy generate an $H$-step, $D$-dimensional action chunk $z\in\mathbb{R}^{H\times D}$, flattened as $z\in\mathbb{R}^{HD}$ when convenient.\par\pagebreak
The flow endpoint denotes the clean generated chunk.

\paragraph{Physical precision.}
Let $z^*$ be the demonstrated chunk, $\hat z$ the generated clean chunk, and $e_z=\hat z-z^*$ the action prediction error.
The map $F_c:\mathbb R^{HD}\to\mathbb R^m$ describes the physical outcome in $m$ task-relevant directions. First-order expansion around $z^*$ gives
\begin{equation}
 \Delta y_c=F_c(z^*+e_z)-F_c(z^*)\approx J_c e_z,
 \qquad J_c=\left.\frac{\partial F_c}{\partial z}\right|_{z^*}.
 \label{eq:physical-map}
\end{equation}
If the admissible errors along $m$ task-space directions are $\delta_1(c),\ldots,\delta_m(c)$, define
\begin{equation}
 \Lambda(c)=\operatorname{diag}\!\left(\delta_1(c)^{-2},\ldots,\delta_m(c)^{-2}\right).
\end{equation}
The local tolerance-normalized endpoint risk is
\begin{equation}
 \mathcal R(c)=\mathbb E\!\left[e_z^\top J_c^\top\Lambda(c)J_c e_z\mid c\right].
 \label{eq:tolerance-risk}
\end{equation}
A tenfold tighter tolerance inflates the squared normalized consequence one hundredfold at unchanged absolute error.
The risk also controls task failure. If failure occurs when $\lVert\Lambda(c)^{1/2}J_ce_z\rVert_2>1$, Markov's inequality applied to the squared norm gives $\Pr(\mathrm{fail}\mid c)\leq\mathcal R(c)$, so Eq.~(\ref{eq:tolerance-risk}) bounds context-conditional failure probability and formalizes the precision demand of a context.

\paragraph{Local flow objective.}
We set $\tau=1$ for noise and $\tau=0$ for clean actions, so generation integrates from $\tau=1$ to $\tau=0$. For $\epsilon\sim\mathcal N(0,I)$,
\begin{equation}
 x_\tau=(1-\tau)z^*+\tau\epsilon,
 \qquad u_\tau=\frac{d x_\tau}{d\tau}=\epsilon-z^*.
 \label{eq:flowpath}
\end{equation}
Let $v_\theta$ be the learned velocity field and $p_\tau$ the flow-time sampling density. The population optimum of squared conditional flow matching is $v^*(x_\tau,\tau,c)=\mathbb E[u_\tau\mid x_\tau,\tau,c]$.
Along an ideal generation trajectory $x^*(\tau)$ driven by $v^*$, define the reducible residual $r_c(\tau)=(v_\theta-v^*)(x^*(\tau),\tau,c)$. Its weighted energy over a flow-time region $B\subseteq[0,1]$ is
\begin{equation}
 \mathcal E_c(B)=\int_B p_\tau(\tau)\lVert r_c(\tau)\rVert_2^2\,d\tau.
 \label{eq:residual-budget}
\end{equation}
This residual energy measures the local error penalized by standard flow matching. Its physical consequence depends on propagation to the endpoint.

\paragraph{From residual to endpoint.}
Linearizing the learned ODE around the ideal trajectory yields
\begin{equation}
 \frac{d\,\delta x}{d\tau}\approx G_c(\tau)\delta x+r_c(\tau),
 \qquad G_c(\tau)=\nabla_x v^*(x^*(\tau),\tau,c).
\end{equation}
Here $\delta x$ is the learned trajectory's deviation from $x^*$, with shared initial noise giving $\delta x(1)=0$, and $G_c$ is the ideal vector-field Jacobian.
The state-transition matrix $\Phi_c(t,s)$ propagates a perturbation from flow time $s$ to $t$. Residuals in $B$ induce the linearized, tolerance-normalized endpoint perturbation
\begin{equation}
 \eta_c(B)=-\int_B A_c(\tau)r_c(\tau)\,d\tau,
 \quad A_c(\tau)=\Lambda(c)^{1/2}J_c\Phi_c(0,\tau).
 \label{eq:propagation}
\end{equation}
The matrix $\Phi$ propagates a local flow error to the clean action, $J$ maps that action error into the physical task space, and $\Lambda$ measures it relative to what the current context tolerates.
Because $\Phi_c(0,\tau)$ depends jointly on flow time and context, and $J_c$ and $\Lambda(c)$ add further context dependence, their product admits no context-independent flow-time weighting in general.

\begin{figure*}[!t]
 \centering
 \includegraphics[width=\linewidth]{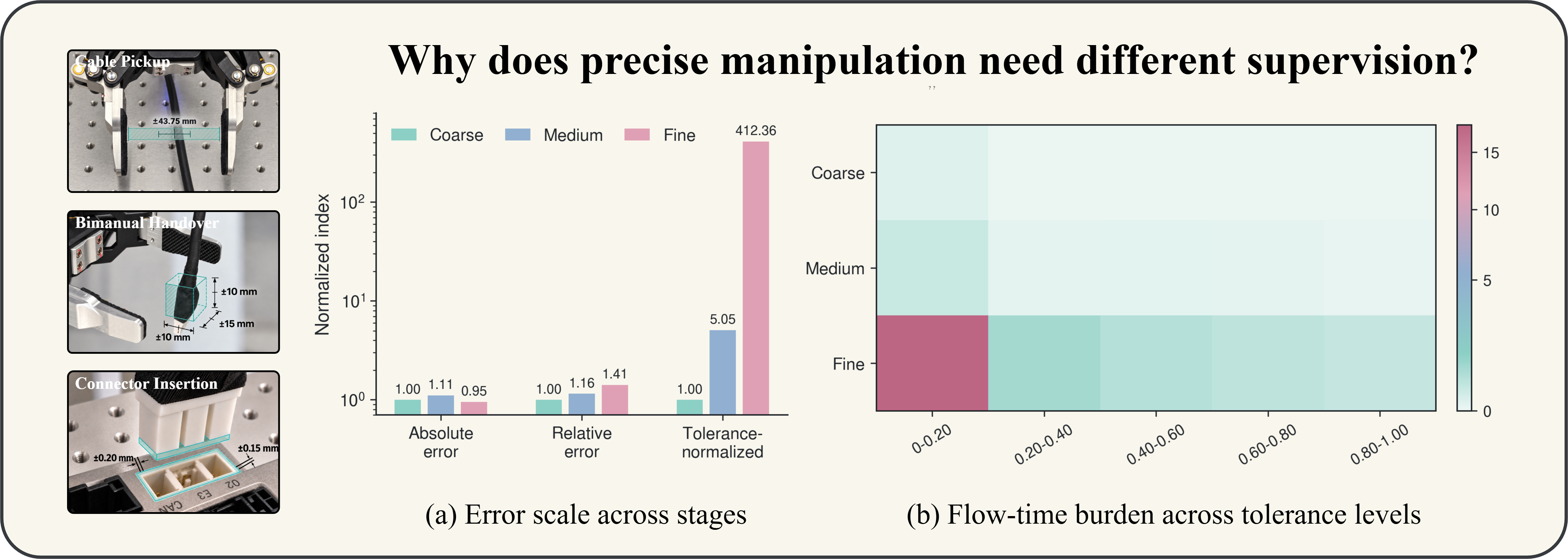}
 \caption{\textbf{Why precise manipulation needs different supervision.} Left, top to bottom: key frames and tolerances for cable pickup, bimanual handover, and connector insertion, the three stages of the real-robot submillimeter drive-insertion task. (a) Similar action errors incur different tolerance-normalized burdens. (b) Inverse-tolerance-weighted endpoint responses to equal-energy velocity perturbations peak near the clean endpoint for fine manipulation.}
 \label{fig:diagnosis}
 \vspace{20pt}
\end{figure*}

\begin{proposition}[Flow-Time Starvation]
For any region $B$ with $p_\tau(\tau)>0$ and square-integrable residual, the linearized tolerance-normalized endpoint error obeys
\begin{equation}
 \lVert\eta_c(B)\rVert_2^2
 \leq \mathcal E_c(B)\,\mathcal S_c(B),
 \quad
 \mathcal S_c(B)=\lambda_{\max}\!\left[
 \int_B\frac{A_c(\tau)A_c(\tau)^\top}{p_\tau(\tau)}\,d\tau
 \right].
 \label{eq:starvation}
\end{equation}
\end{proposition}
Here $\lambda_{\max}$ denotes the largest eigenvalue, and $\mathcal S_c(B)$ measures worst-case endpoint sensitivity per unit residual energy.
\noindent\textit{Derivation.}
For any unit task-space direction $h$, multiplying and dividing the integrand $h^\top A_c(\tau)r_c(\tau)$ by $\sqrt{p_\tau(\tau)}$ and applying weighted Cauchy Schwarz bounds its square by $\mathcal E_c(B)$ times
$h^\top[\int_B A_cA_c^\top/p_\tau]h$, and maximizing over $\lVert h\rVert_2=1$ gives Eq.~(\ref{eq:starvation}) through the Rayleigh quotient.
Hence $\mathcal S_c(B)$ is the squared operator norm from the weighted local-residual space to the tolerance-normalized task space, so one unit of standard FM penalty can leave a large endpoint consequence and the region is starved relative to its physical leverage.
The bound is tight for residuals proportional to $p_\tau(\tau)^{-1}A_c(\tau)^\top h_{\max}$, where $h_{\max}$ is a leading eigenvector of the integrated matrix.

\paragraph{Visible supervision.}
The bound identifies the ideal allocation through $\Lambda$, $J$, $\Phi$, and $p_\tau$, but the first three are unavailable from demonstrations, so a practical objective needs observable surrogates that also avoid amplifying a sample before its action signal becomes visible in the noisy interpolation.
For the normalized ground-truth chunk, we therefore define the sample-specific signal-to-noise ratio (SNR) as
\begin{equation}
 \sigma_{f,i}=\sqrt{\frac{1}{HD}\sum_{h=1}^{H}\sum_{d=1}^{D}\left(z^*_{i,h,d}\right)^2},
 \qquad
 \operatorname{SNR}_i=\frac{(1-\tau_i)^2\sigma_{f,i}^2}{\tau_i^2+\varepsilon}.
 \label{eq:snr}
\end{equation}
Here $z^*_{i,h,d}$ is the normalized action component at step $h$ and dimension $d$ of sample $i$, $\sigma_{f,i}$ is its chunk RMS scale, $\tau_i$ is the sampled flow time, and $\varepsilon>0$ stabilizes the denominator.

The gate has a simple probabilistic interpretation.
Assume independent action-evidence arrivals with count
$N_i\sim\operatorname{Poisson}(\operatorname{SNR}_i/\tau_0)$, where $\tau_0>0$ controls the visibility saturation scale.
Then the probability of observing at least one useful event is
\begin{equation}
 \Pr(N_i\geq1)=1-\Pr(N_i=0)=1-\exp(-\operatorname{SNR}_i/\tau_0).
 \label{eq:poisson-gate}
\end{equation}
The Poisson assumption converts the SNR into a saturating visibility probability, with Gaussian interpolation setting the SNR and the arrival model gating it.

\paragraph{Empirical diagnosis.}
Figure~\ref{fig:diagnosis} diagnoses this mismatch on drive connector insertion, with grasping, handover, and insertion tolerances of $43.75$, $10$, and $0.15$\,mm. Absolute FM errors remain comparable, but tolerance-normalized burden rises as tolerance narrows (a). Panel (b) weights endpoint action MSE per unit velocity-perturbation energy by each stage's inverse tolerance relative to grasping. This proxy response peaks for fine manipulation near the clean endpoint (see \textit{Supplementary Material} for the protocol and unweighted responses). These patterns motivate context-dependent precision scoring and clean-end supervision in Section~\ref{sec:prism}.

\begin{figure*}[!t]
 \centering
 \includegraphics[width=\linewidth]{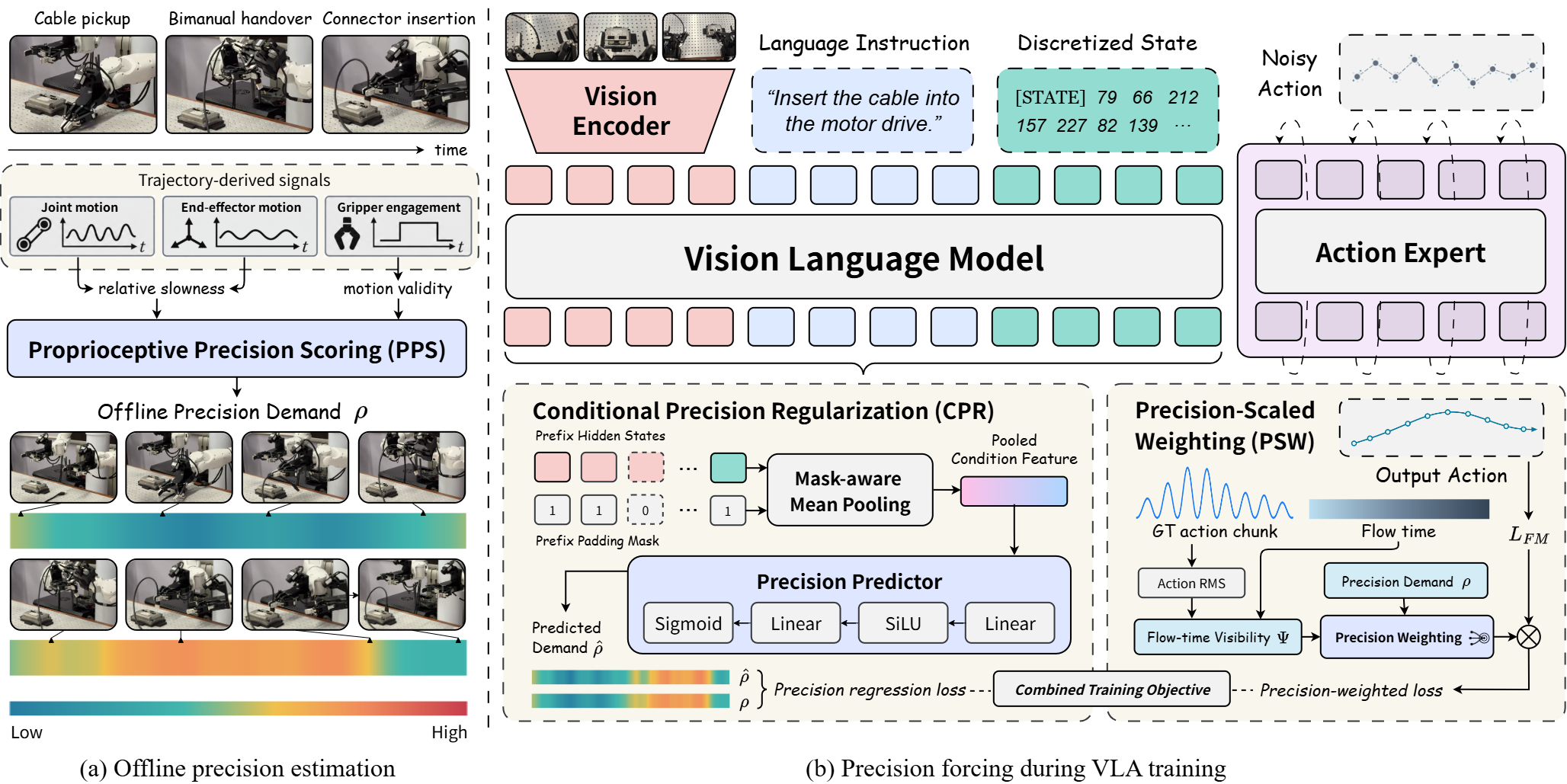}
 \caption{\textbf{Overview of \method{}.} (a) PPS combines joint motion, end-effector motion, and gripper engagement into offline precision demand. (b) CPR predicts this demand from mask-pooled VLM prefix features, while PSW reweights the flow-matching loss using offline demand and flow-time visibility derived from action RMS and flow time. Both objectives leave inference unchanged.}
 \label{fig:architecture}
\end{figure*}

\section{PRISM}
\label{sec:prism}

\subsection{Overview}
\method{} uses a shared precision signal to shape both action supervision and the VLM condition representation.
PPS computes a frame-level precision-demand label $\rho_i\in[0,1]$ offline and fixes it before optimization.
PSW uses this label to allocate more action supervision to precision-demanding samples, while CPR trains the condition representation to encode the same demand.
Both objectives use the offline label directly, and the CPR prediction does not feed back into sample weighting.
The auxiliary labels and prediction head are discarded after training, leaving the backbone and its standard ODE solver unchanged at inference.

\subsection{Proprioceptive Precision Scoring}
PPS estimates $\rho_i$ from demonstrated joint displacement, end-effector displacement, and gripper state.
Because demonstrations record robot motion rather than the latent geometry and contact state that determine physical tolerance, PPS combines motion, engagement, acquisition, and quiescence cues that recur during precision-demanding behavior.
For frame $i$, define the motion measurements and their normalized mixture as follows.
\begin{equation}
 \Delta q_i=\lVert q_i-q_{i-1}\rVert_2,\quad
 \Delta e_i=\max_{b\in\{L,R\}}\lVert e_i^b-e_{i-1}^b\rVert_2,
 \quad s_i=\alpha_q\Delta q_i+\alpha_e\Delta e_i.
 \label{eq:pps-motion}
\end{equation}
Here $q_i$ and $e_i^b$ are the joint and arm-$b$ end-effector states. The nonnegative mixture coefficients satisfy $\alpha_q+\alpha_e=1$.
Within active frames of the episode, percentile rank converts motion into
$\operatorname{slow}_i=1-\operatorname{rank}_{\mathrm{episode}}(s_i)$.
Let $g_i^{\mathrm{eng}}\in[0,1]$ be normalized gripper engagement and let $t_i\in[0,1]$ combine an acquisition gate, which suppresses rapid closing, and a quiescent-hand gate, which suppresses isolated gripper movement while the arms are static.
We set
\begin{equation}
 \rho_i=g_i^{\mathrm{eng}}\operatorname{slow}_i t_i,
 \label{eq:pps}
\end{equation}
then apply fixed neutral values to disengaged and near-pause frames, temporal smoothing, and clipping to $[0,1]$ as specified in Section~\ref{sec:implementation-details}.
Each anchor frame contributes one label to its action chunk, so PSW does not average or maximize future $\rho$ values within the chunk.
Section~\ref{sec:ablations} evaluates whether this offline behavioral proxy tracks physical precision demand across tasks.

\subsection{Precision-Scaled Weighting}
Section~\ref{sec:problem_formulation} characterizes worst-case endpoint sensitivity through $\mathcal S_c(B)$. Its underlying context factor $\Lambda(c)^{1/2}J_c$ and propagation factor $\Phi_c(0,\tau)$ are not directly available during training.
PSW approximates these dependencies with a separable surrogate. The precision score $\rho_i$ represents context-dependent demand, a monotone function of $\tau$ emphasizes the clean end, and the visibility probability in Eq.~(\ref{eq:poisson-gate}) delays amplification until the noisy interpolation contains usable action evidence.
Figure~\ref{fig:diagnosis}(b) motivates this separable design through a tolerance-weighted response that combines tighter task constraints with stronger clean-end perturbation effects.
\begin{equation}
 \Psi_i=(1-\tau_i)^p\left(1-e^{-\operatorname{SNR}_i/\tau_0}\right),
 \qquad w_i=1+\gamma\rho_i\Psi_i.
 \label{eq:psw}
\end{equation}
The positive hyperparameters $p$, $\tau_0$, and $\gamma$ control clean-end emphasis, visibility saturation, and maximum amplification, respectively.
Since $\Psi_i\in[0,1]$, the weight satisfies $w_i\in[1,1+\gamma]$, so PSW only adds supervision to precision-demanding samples and never withdraws supervision from any other sample.
The reweighted flow-matching objective for sample $i$ is
\begin{equation}
 \mathcal L_{\mathrm{PSW},i}
 =\frac{w_i}{HD}\sum_{h=1}^{H}\sum_{d=1}^{D}
 \left(v_{\theta,i,h,d}-u_{i,h,d}\right)^2.
 \label{eq:pswloss}
\end{equation}
The weight multiplies the chunk as a whole, so PSW redistributes gradient magnitude across samples and flow times rather than across action dimensions.
Flow time alone does not determine action visibility. At the same $\tau$, a small-amplitude precision chunk can have a lower SNR than a larger routine chunk, causing a flow-time-only rule to amplify supervision while the target remains dominated by interpolation noise.
The saturating gate delays this amplification until usable action evidence appears, and Section~\ref{sec:ablations} evaluates its incremental effect after clean-end weighting.
PSW reallocates gradients without changing the flow-time sampler, noise schedule, or number of action experts.

\subsection{Conditional Precision Regularization}
PSW allocates action supervision according to precision demand, while CPR trains the condition representation to encode that demand.
This representation-level supervision helps the policy recognize when greater precision is needed.
For the mask-aware pooled VLM condition representation $h_{c_i}$, a lightweight multilayer perceptron (MLP) head predicts
\begin{equation}
 \hat\rho(c_i)=\sigma(\operatorname{MLP}(h_{c_i})),
 \qquad \mathcal L_{\mathrm{CPR},i}=(\hat\rho(c_i)-\rho_i)^2.
 \label{eq:cpr}
\end{equation}
Jointly training the head and conditioning backbone allows the precision target to shape the representation itself.
PSW continues to use the offline target $\rho_i$ rather than $\hat\rho$, keeping sample weighting independent of the model's current prediction.
The complete objective reweights the flow-matching loss and adds exactly one auxiliary term,
\begin{equation}
 \mathcal L=\mathcal L_{\mathrm{PSW}}+\lambda_{\mathrm{CPR}}\mathcal L_{\mathrm{CPR}}.
\end{equation}
Beyond the fixed PPS settings, \method{} therefore introduces four scalar hyperparameters, namely $p$, $\tau_0$, $\gamma$, and $\lambda_{\mathrm{CPR}}$, and no inference-time module.

\section{Experiments}
\label{sec:experiments}

\begin{table*}[t]
  \centering
  \caption{RLBench-10 success rates (absolute gains in parentheses). Green highlights \method{}.}
  \label{tab:rlbench10_literature}
  \small
  \setlength{\tabcolsep}{3.1pt}
  \renewcommand{\arraystretch}{0.88}
  \resizebox{\textwidth}{!}{%
  \begin{tabular}{lccccccccccc}
    \toprule
    \rowcolor{simtableheader}
    \rule[-0.8ex]{0pt}{3.0ex}Method & Close box & Close laptop & Toilet seat & Sweep & Close fridge & Phone & Umbrella & Frame & Wine & Water plants & Mean \\
    \midrule
    OpenVLA      & 0.60 & 0.35 & 0.75 & 0.55 & 0.85 & 0.20 & 0.30 & 0.15 & 0.20 & 0.05 & 0.40 \\
    SpatialVLA   & 0.80 & 0.70 & 0.85 & 0.20 & 0.80 & 0.15 & 0.25 & 0.40 & 0.15 & 0.30 & 0.46 \\
    $\pi_0$       & 0.90 & 0.60 & \textbf{1.00} & 0.30 & 0.90 & 0.25 & 0.35 & 0.75 & 0.05 & 0.45 & 0.55 \\
    CogACT       & 0.90 & 0.80 & 0.95 & 0.50 & 0.85 & 0.50 & 0.55 & 0.45 & 0.30 & 0.25 & 0.61 \\
    CoT-VLA      & \textbf{0.95} & 0.75 & \textbf{1.00} & 0.80 & 0.65 & 0.50 & 0.40 & 0.50 & 0.55 & 0.50 & 0.66 \\
    HybridVLA    & 0.85 & 0.95 & \textbf{1.00} & \textbf{0.90} & \textbf{1.00} & 0.50 & 0.50 & 0.70 & 0.50 & 0.50 & 0.74 \\
    LaST$_0$     & \textbf{0.95} & 0.95 & \textbf{1.00} & 0.80 & 0.85 & 0.75 & 0.75 & 0.70 & 0.85 & 0.60 & 0.82 \\
    \midrule
    StarVLA & 0.80 & 0.95 & 0.85 & 0.75 & 0.90 & 0.45 & 0.60 & 0.60 & 0.85 & 0.45 & 0.72 \\
    \rowcolor{simtableprism}
    StarVLA + \method{} & 0.85 & 0.95 & 0.95 & 0.85 & 0.95 & 0.60 & 0.70 & 0.75 & \textbf{1.00} & 0.60 & 0.82 {\scriptsize(+0.10)} \\
    $\pi_{0.5}$ & 0.90 & 0.95 & 0.85 & 0.80 & \textbf{1.00} & 0.55 & 0.60 & 0.70 & 0.80 & 0.45 & 0.76 \\
    \rowcolor{simtableprism}
    $\pi_{0.5}$ + \method{} & 0.90 & \textbf{1.00} & \textbf{1.00} & \textbf{0.90} & \textbf{1.00} & \textbf{0.85} & \textbf{0.80} & \textbf{0.80} & 0.90 & \textbf{0.65} & \textbf{0.88} {\scriptsize(+0.12)} \\
    \bottomrule
  \end{tabular}%
  }
\end{table*}

\subsection{Experimental Setup}
\paragraph{Backbones.}
We evaluate \method{} with $\pi_{0.5}$ and the $\pi$-style flow-matching variant of StarVLA~\citep{starvla} on simulation benchmarks.
The real-robot study uses $\pi_{0.5}$.
Each controlled comparison shares the observations, pretrained initialization, demonstrations, update budget, and inference schedule.

\paragraph{Simulation.}
RLBench-10 evaluates one multi-task policy across ten language-conditioned tasks, while LIBERO contains the Spatial, Object, Goal, and Long suites~\citep{rlbench,libero}.
We report rollout success by task or suite and its mean.
RLBench baselines include OpenVLA, SpatialVLA, $\pi_0$, CogACT, CoT-VLA, HybridVLA, and LaST$_0$~\citep{openvla,spatialvla,pi0,cogact,cotvla,hybridvla,last0}.
The LIBERO comparison includes SpatialVLA, CoT-VLA, UnifiedVLA, GR00T N1, OpenVLA-OFT, and $\pi_0$~\citep{spatialvla,cotvla,unifiedvla,grootn1,openvlaoft,pi0}.
Controlled results use each backbone with and without \method{}.

\paragraph{Real robot.}
Multi-Stage Submillimeter-Tolerance Manipulation (MSSM) evaluates extended bimanual tasks with precision-critical alignment and contact stages. Drive connector insertion has a measured geometric clearance of $0.15$\,mm.
We evaluate three MSSM tasks on the Marvin dual-arm robot using \mssmdemos{} demonstrations per task.
Each method is evaluated in 20 trials per task. Both end-to-end and individual-stage success rates use a denominator of 20.
We report mean completion time over successful trials. A precision stage fails when the robot releases the object or tool after unsuccessful insertion or tightening. Any stage failure makes the full trial unsuccessful. Bimanual mating continues to later connector pairs after an earlier failure, so stage success need not decrease monotonically.
We compare $\pi_{0.5}$, its DAgger variant trained with \daggerdemos{} additional corrective demonstrations per task, and \method{} applied to the same base policy.

\begin{figure*}[!t]
 \centering
 \includegraphics[width=\linewidth]{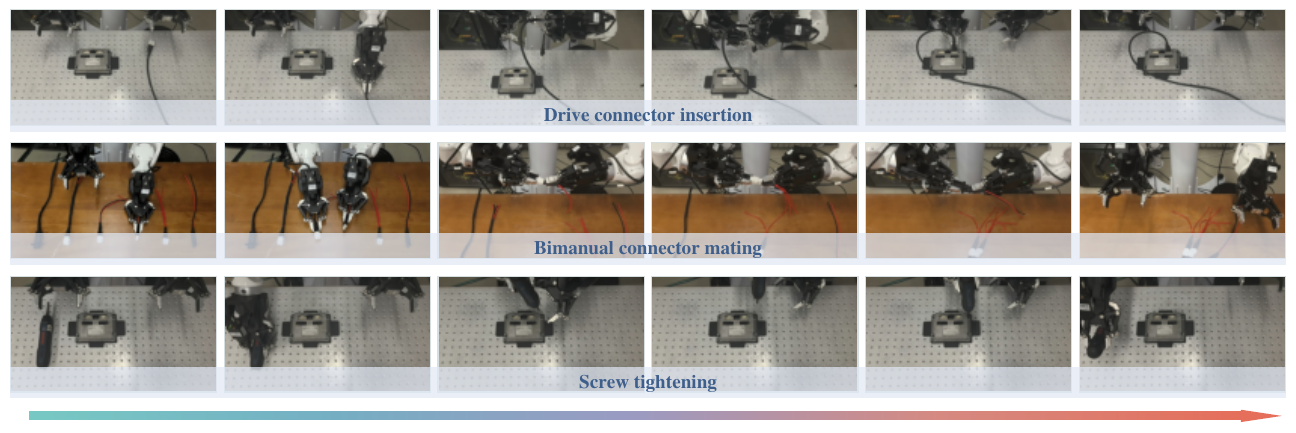}
 \caption{Representative stages of three MSSM tasks, ordered temporally from left to right.}
 \label{fig:mssm-execution-progress}
\end{figure*}

\subsection{Implementation Details}
\label{sec:implementation-details}
\begingroup
\parshape=16
0pt 0.44\textwidth
0pt 0.44\textwidth
0pt 0.44\textwidth
0pt 0.44\textwidth
0pt 0.44\textwidth
0pt 0.44\textwidth
0pt 0.44\textwidth
0pt 0.44\textwidth
0pt 0.44\textwidth
0pt 0.44\textwidth
0pt 0.44\textwidth
0pt 0.44\textwidth
0pt 0.44\textwidth
0pt 0.44\textwidth
0pt 0.44\textwidth
0pt \textwidth
\noindent
\makebox[0pt][l]{%
 \hspace{0.47\textwidth}%
 \raisebox{0pt}[0pt][0pt]{%
  \begin{minipage}[t]{0.53\textwidth}
   \vspace{0pt}
   \centering
   \captionof{table}{LIBERO success rates (\%) with absolute gains. \method{} models are highlighted in green.}
   \label{tab:libero}
   \scriptsize
   \setlength{\tabcolsep}{2.4pt}
   \renewcommand{\arraystretch}{0.80}
   \resizebox{\linewidth}{!}{%
   \begin{tabular}{lccccc}
    \toprule
    \rowcolor{simtableheader}
    \rule[-0.8ex]{0pt}{3.0ex}Method & Spatial & Object & Goal & Long & Avg. \\
    \midrule
    SpatialVLA                & 88.2 & 89.9 & 78.6 & 55.5 & 78.1 \\
    CoT-VLA                   & 87.5 & 91.6 & 87.6 & 69.0 & 81.1 \\
    UnifiedVLA                & 96.5 & 96.8 & 95.6 & 92.0 & 95.2 \\
    GR00T N1                  & 94.4 & 97.6 & 93.0 & 90.6 & 93.9 \\
    $\pi_0$                   & 96.8 & 98.8 & 95.8 & 85.2 & 94.2 \\
    OpenVLA-OFT               & 97.6 & 98.4 & 97.9 & 94.5 & 97.1 \\
    \midrule
    StarVLA                   & 98.8 & \textbf{99.6} & 95.8 & 88.4 & 95.7 \\
    \rowcolor{simtableprism}
    StarVLA + \method{}       & \textbf{99.0} & 99.4 & 97.2 & 90.2 & 96.5 {\scriptsize(+0.8)} \\
    $\pi_{0.5}$               & 98.8 & 98.2 & 98.0 & 92.4 & 96.9 \\
    \rowcolor{simtableprism}
    $\pi_{0.5}$ + \method{}  & 98.6 & 99.4 & \textbf{98.2} & 93.8 & \textbf{97.5} {\scriptsize(+0.6)} \\
    \bottomrule
   \end{tabular}%
   }
  \end{minipage}%
 }}%
For $\pi_{0.5}$ experiments, we use an action horizon of 15, and simulation datasets use an action stride of 1 at their stored control rate. We optimize $\pi_{0.5}$ for 15,000 steps with AdamW, a global batch size of 192, gradient clipping at 1, an exponential moving average decay of $0.999$, and 500 warmup steps followed by a constant learning rate of $5\times10^{-5}$. Training uses eight NVIDIA H20 GPUs with data parallelism and an FSDP axis of one. The real-robot connector-insertion configuration uses $p=2$, $\tau_0=1$, $\gamma=3$, and $\lambda_{\mathrm{CPR}}=0.1$. PPS combines joint and end-effector motion with equal weights, uses a disengaged baseline of $0.35$ and an inactive-frame score of $0.5$, and applies seven-frame smoothing. Paired baseline and \method{} runs share initialization, data order, update budget, and inference settings.
Real-robot experiments use the Marvin dual-arm platform with HIKROBOT MV-CS016-10GC cameras. Demonstrations are collected through teleoperation using a DPVR E4 Black headset and controllers.
\par
\endgroup

\subsection{Simulation Results}
\label{sec:simulation-results}
RLBench-10 contains more precision-sensitive interactions, where \method{} raises mean success by $0.10$ for StarVLA and $0.12$ for $\pi_{0.5}$, demonstrating large gains when fine manipulation determines success. LIBERO is simpler and near saturation, yet \method{} still improves the averages by $0.8$ and $0.6$ points, respectively, showing that the benefit extends beyond precision-heavy benchmarks. We hypothesize that precision-aware training better distinguishes coarse-motion and fine-adjustment regions of the action space, benefiting general manipulation as well.

\subsection{MSSM Real-Robot Results}
\label{sec:real-results}
\begin{table*}[t]
 \centering
 \caption{MSSM real-robot results. S1--S5 report stage success. SR is end-to-end task success. $T$ is mean completion time in seconds over successful trials. Avg.\ averages SR over the three tasks.}
 \label{tab:real-robot-main}
 \small
 \setlength{\tabcolsep}{2.3pt}
 \renewcommand{\arraystretch}{0.88}
 \resizebox{\textwidth}{!}{%
 \begin{tabular}{@{}l*{5}{c}@{\hspace{6pt}}*{7}{c}@{\hspace{6pt}}*{5}{c}@{\hspace{6pt}}c@{}}
  \toprule
  \rowcolor{simtableheader}
  \rule[-1.1ex]{0pt}{3.8ex}Method &
  \multicolumn{5}{c}{\shortstack{Drive connector insertion\\(0.15 mm)}} &
  \multicolumn{7}{c}{\shortstack{Bimanual connector mating\\(0.1 mm)}} &
  \multicolumn{5}{c}{\shortstack{Screw tightening\\(0.2 mm)}} & Avg. \\
  \cmidrule(lr){2-6}\cmidrule(lr){7-13}\cmidrule(lr){14-18}
  & S1 & S2 & S3 & SR$\uparrow$ & $T\downarrow$ & S1 & S2 & S3 & S4 & S5 & SR$\uparrow$ & $T\downarrow$ & S1 & S2 & S3 & SR$\uparrow$ & $T\downarrow$ & SR$\uparrow$ \\
  \midrule
  $\pi_{0.5}$ & 1.00&0.90&0.35&0.35&47 & 0.95&0.95&0.50&0.55&0.65&0.35&65 & 0.95&0.85&0.45&0.45&37 & 0.38 \\
  $\pi_{0.5}$ + DAgger & 1.00&1.00&0.65&0.65&43 & 0.95&1.00&0.70&0.75&0.85&0.60&59 & 1.00&0.95&0.70&0.70&35 & 0.65 \\
  \rowcolor{simtableprism}
  $\pi_{0.5}$ + \method{} & 1.00&1.00&0.95&\textbf{0.95}&\textbf{38} & 1.00&1.00&0.85&0.90&0.95&\textbf{0.80}&\textbf{52} & 1.00&1.00&0.85&\textbf{0.85}&\textbf{32} & \textbf{0.87} \\
 \bottomrule
 \end{tabular}
 }
\end{table*}

\noindent
Across drive connector insertion, bimanual connector mating, and screw tightening (Table~\ref{tab:real-robot-main}), \method{} raises end-to-end success from $0.35/0.35/0.45$ to $0.95/0.80/0.85$, versus $0.65/0.60/0.70$ after DAgger. Mean task success increases from $0.38$ to $0.87$, while successful-trial times decrease from $47/65/37$ to $38/52/32$\,s. We observed that DAgger's advantage was less apparent with varied placements or unfavorable pre-alignment configurations. \method{} preserved reliable coarse motions while making more directed local corrections, whereas the baselines more often repeated attempts without resolving misalignment. This suggests improved recovery during precision-critical interaction.

\subsection{Ablation Studies}
\label{sec:ablations}
We cumulatively add CPR, precision weighting ($+\rho$), clean-end emphasis ($+\tau$), and SNR gating on drive insertion (Table~\ref{tab:real-ablation}).
One attempt is counted each time the robot applies force to insert the connector after an alignment motion.
\begin{table*}[!t]
 \centering
 \begin{minipage}[t]{0.54\linewidth}
 \captionsetup{font=small}
 \caption{Left-to-right cumulative ablation on drive insertion. Time and attempts average successful trials. $+\mathrm{SNR}$ is full \method{}. Arrows indicate better performance.}
 \label{tab:real-ablation}
 \footnotesize
 \setlength{\tabcolsep}{3pt}
 \renewcommand{\arraystretch}{0.94}
 \begin{tabular*}{\linewidth}{@{\extracolsep{\fill}}lccccc@{}}
  \toprule
  Metric & $\pi_{0.5}$ & $+\,$CPR & $+\,\rho$ & $+\,\tau$ & \textbf{$\boldsymbol{+}$ SNR} \\
  \midrule
  Success rate $\uparrow$ & 0.35 & 0.45 & 0.70 & 0.85 & \textbf{0.95} \\
  Time (s) $\downarrow$ & 47 & 45 & 43 & 40 & \textbf{38} \\
  Attempts $\downarrow$ & 4.1 & 3.5 & 2.7 & 2.4 & \textbf{2.1} \\
  \bottomrule
 \end{tabular*}
 \end{minipage}\hfill
 \begin{minipage}[t]{0.44\linewidth}
 \captionsetup{font=small}
 \caption{State-conditioning ablation on drive insertion. Base and \method{} success rates. $\pi_0$ retains proprioceptive input to its action expert.}
 \label{tab:state-conditioning}
 \footnotesize
 \setlength{\tabcolsep}{3pt}
 \renewcommand{\arraystretch}{0.94}
 \begin{tabular*}{\linewidth}{@{\extracolsep{\fill}}lccc@{}}
  \toprule
  Backbone & VLM state & Base & +\method{} \\
  \midrule
  $\pi_{0.5}$ & Yes & 0.35 & \textbf{0.95} \\
  $\pi_{0.5}$ & No & 0.30 & \textbf{0.95} \\
  $\pi_0$ & No & 0.25 & \textbf{0.80} \\
  \bottomrule
 \end{tabular*}
 \end{minipage}
\end{table*}
Success rises from $0.35$ to $0.45$ with CPR and $0.70$ with precision weighting, then to $0.85$ and $0.95$ with clean-end emphasis and SNR gating. We observe more agile fine motions with clean-end emphasis and stronger local adjustment with SNR gating. Attempts decrease from $4.1$ to $2.1$ across the cumulative design.

\subsection{Discussion}

\begin{figure*}[!t]
  \centering
  \begin{minipage}[b]{0.262\textwidth}
    \centering
    \includegraphics[width=\linewidth]{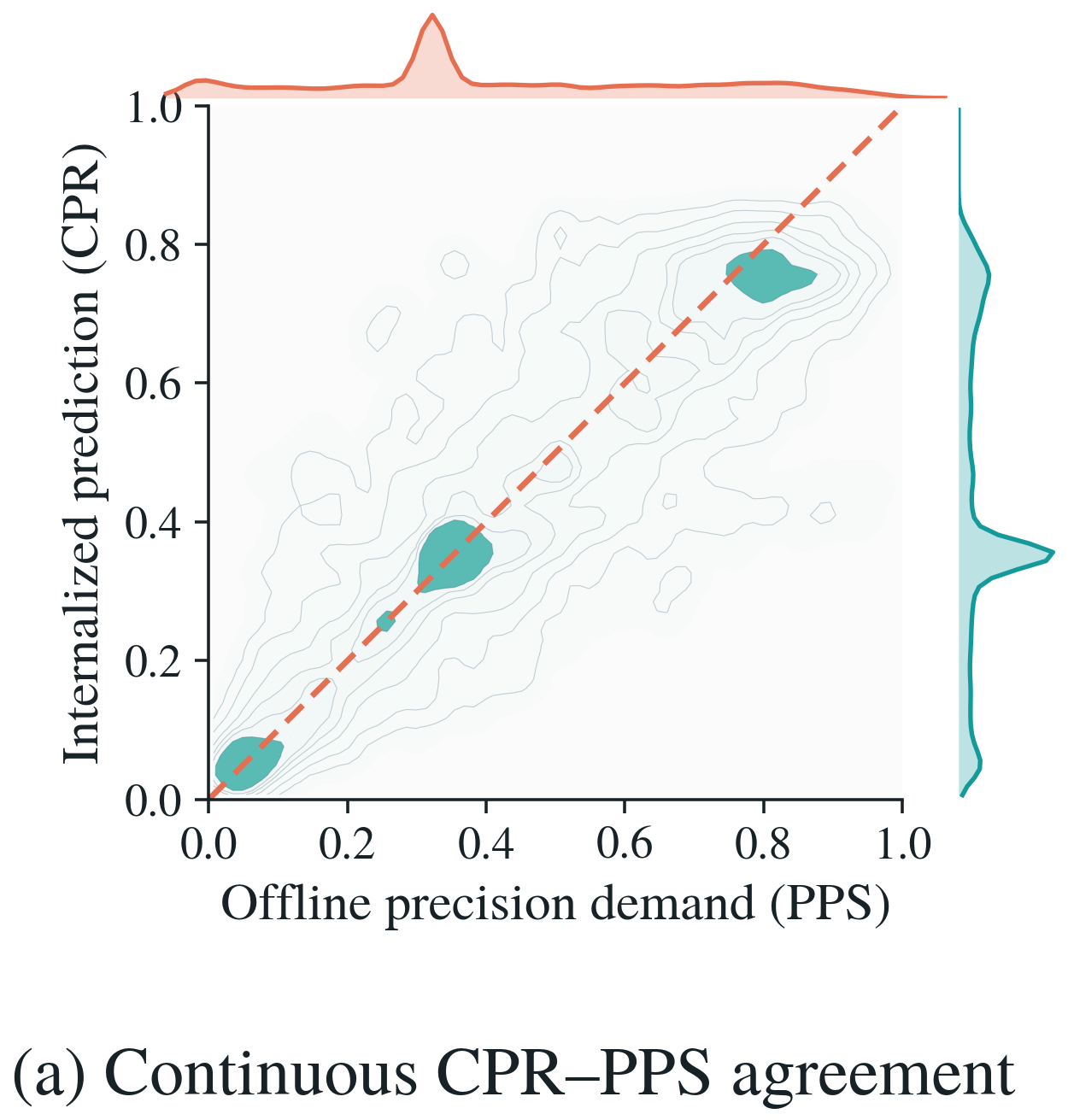}
  \end{minipage}
  \hfill
  \begin{minipage}[b]{0.718\textwidth}
    \centering
    \includegraphics[width=\linewidth]{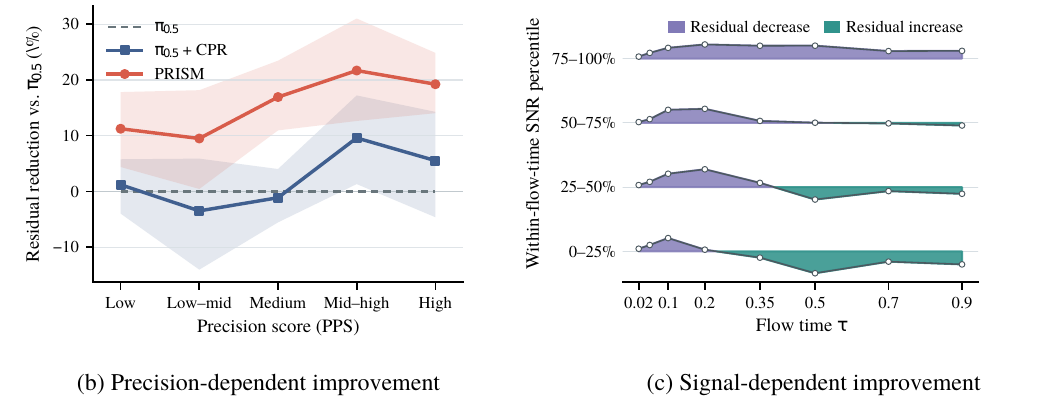}
  \end{minipage}
  \par\vspace{7pt}
  \includegraphics[width=0.99\linewidth]{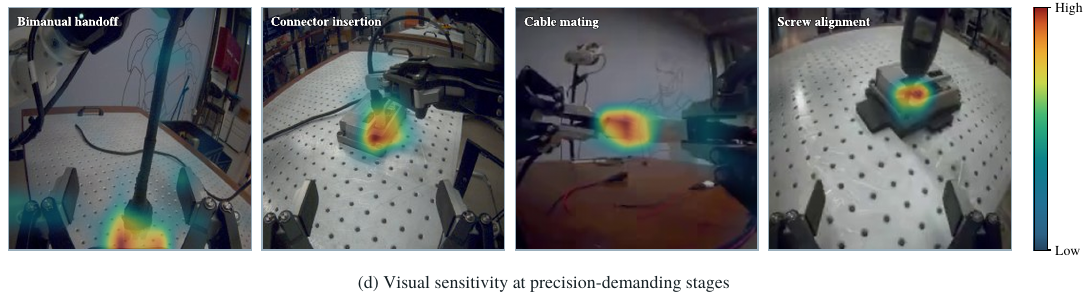}
  \caption{\textbf{Mechanism analysis of precision forcing.} (a) CPR predictions versus offline PPS on held-out insertion trajectories. (b) Residual reduction over $\pi_{0.5}$ across five equal-count PPS bins. (c) Residual profiles in four equal-count SNR groups, formed per flow time within high-PPS frames. Purple/teal indicates a decrease/increase under PSW. (d) Relative CPR-conditioned visual sensitivity across fine-manipulation contexts.}
  \label{fig:mechanism_discussion}
\end{figure*}

\noindent\textbf{PPS estimation accuracy and generality.}
On 6,692 frames from ten held-out insertion episodes, CPR predicts offline PPS with Spearman/Pearson correlations of $0.915/0.931$ and mean absolute error $0.056$ (Figure~\ref{fig:mechanism_discussion}(a)), preserving its continuous demand structure.
Across nineteen tasks, PPS highlights fine-operation intervals in mixed coarse--fine trajectories and assigns low demand to coarse-only tasks, supporting cross-task applicability. Task descriptions, PPS computation details, and the corresponding trajectory visualizations are provided in the \textit{Supplementary Material}.

\noindent\textbf{Complementary roles of CPR and PSW.}
CPR lowers overall residuals but worsens two lower-demand bins, suggesting that recognizing precision demand alone does not ensure balanced action learning. Adding PSW improves every bin (Figure~\ref{fig:mechanism_discussion}(b)). At fixed $\tau$, gains favor higher-SNR groups, with little or negative change at low SNR (Figure~\ref{fig:mechanism_discussion}(c)). These patterns support complementary roles: CPR identifies precision-demanding contexts, while PSW allocates supervision according to both precision demand and action visibility.

\noindent\textbf{Recognizing precision-critical contexts.}
Figure~\ref{fig:mechanism_discussion}(d) shows operation-specific CPR sensitivity across four contexts. During handoff, responses concentrate around the held connector and interacting grippers. Drive insertion and connector mating emphasize the mating interfaces, while screw tightening elicits responses near the tool--screw contact region. The relevant regions thus vary with the operation, suggesting that the condition representation captures local visual cues for precision demand rather than a fixed scene location.

\noindent\textbf{State-conditioning ablation.}
We remove state inputs from the $\pi_{0.5}$ VLM during both training and inference (Table~\ref{tab:state-conditioning}): baseline insertion success changes from $0.35$ to $0.30$, while \method{} remains at $0.95$. For further validation, we use $\pi_0$, which feeds state to the action expert. \method{} raises success from $0.25$ to $0.80$. Both settings support CPR learning precision demand from visual-language features without requiring VLM state inputs.

\section{Conclusion}
Generalist VLA policies struggle when brief, precision-critical interactions receive insufficient supervision. We address this problem with \method{}, coupling trajectory-derived precision demand with flow-matching loss weighting and precision-aware representation learning. Simulation and real-robot experiments show improved fine manipulation while preserving general-task performance and leaving inference unchanged. A limitation is the two-stage workflow: PPS targets are computed offline before policy training. Learning precision supervision jointly with the policy, without a separate scoring stage, remains future work. We hope \method{} will serve as a solid baseline for extending VLA policies to precision manipulation.

\bibliography{iclr2027_conference}
\bibliographystyle{iclr2027_conference}

\clearpage
\appendix
\begin{center}
{\large\bfseries Supplementary Material for\par}
\vspace{5pt}
\resizebox{\linewidth}{!}{\bfseries PRISM: Precision Forcing for Generalist Vision-Language-Action Policies}
\end{center}
\vspace{4pt}
\section{Related Work}
\label{sec:related_work}

\paragraph{Vision-language-action policies.}
VLA policies translate visual and language representations into robot actions through token prediction or continuous action generation. RT-1, RT-2, and OpenVLA establish token-based policies trained across diverse tasks~\citep{rt1,rt2,openvla}. SpatialVLA enriches spatial representations, while CogACT, CoT-VLA, and HybridVLA explore the coupling between multimodal reasoning and action prediction~\citep{spatialvla,cogact,cotvla,hybridvla}. These developments improve the representations and decoding mechanisms available to generalist policies.

Our work focuses on policies that couple a VLM to an iterative action expert. $\pi_0$ uses flow matching for cross-embodiment control, and $\pi_{0.5}$ extends this direction toward open-world generalization~\citep{pi0,pi05}. GR00T N1 and SmolVLA also pair multimodal conditioning with generative action prediction~\citep{grootn1,smolvla}. Their action objectives supervise denoising or vector-field prediction locally. PRISM distributes this supervision across contexts with different precision requirements without altering inference.

\paragraph{Precision-sensitive manipulation.}
Precision-sensitive manipulation benefits from contact information and local corrections. ForceVLA, TA-VLA, and Tactile-VLA incorporate force, torque, or tactile signals into manipulation policies~\citep{forcevla,tavla,tactilevla}. Reactive Diffusion Policy combines visual planning with fast tactile corrections, while From Reach to Insert separates visual approach from tactile-augmented reinforcement learning for insertion~\citep{rdp,reachtoinsert}. These approaches make contact dynamics more observable or introduce control mechanisms dedicated to the final interaction.

Demonstration structure offers another route to precision. ACT uses action chunking for fine-grained bimanual behavior, and JUICER combines corrective demonstrations with data augmentation for assembly~\citep{act,juicer}. PRISM instead derives precision demand from the proprioceptive trajectories already present in demonstrations. It uses that demand both to allocate action supervision and to train the condition representation, complementing improvements in sensing, control, and data collection.

\paragraph{Diffusion and flow training objectives.}
Diffusion Policy models action distributions through iterative denoising~\citep{diffusionpolicy}, while flow matching trains vector fields along probability paths that are integrated to generate samples~\citep{flowmatching}. Both objectives depend on supervision at intermediate noise levels. In image generation, Perception Prioritized training reweights denoising losses according to SNR, and Min-SNR weighting reduces conflicts among diffusion timesteps~\citep{p2,minsnr}. These methods establish that uniform treatment of noise levels need not yield the most useful learning signal.

Robot manipulation adds a physical source of imbalance: the same action residual can have different consequences under different tolerances, and local errors can propagate differently to the executed action endpoint. Flow-Time Starvation characterizes this mismatch between local supervision and endpoint burden. PSW couples context-dependent precision demand with clean-end emphasis and SNR-based visibility, so its weights depend on both the manipulation and the generative state. CPR predicts precision demand from the VLM condition representation to complement this allocation.

\section{Training and PPS Implementation Details}
\label{app:pps-spec}
Table~\ref{tab:implementation-settings} expands Section~\ref{sec:implementation-details} with backbone-specific training configurations. StarVLA uses $8\times12\times2=192$ samples per update with gradient accumulation, while $\pi_{0.5}$ uses $8\times24=192$. The real-robot study uses $\pi_{0.5}$ and \mssmdemos{} demonstrations per MSSM task. The DAgger comparison adds \daggerdemos{} corrective demonstrations per task in a second stage. Real-robot control runs at 15\,Hz with a 15-action prediction horizon and replans after executing 12 actions, nominally every 0.8\,s.

\paragraph{PPS computation.}
The implementation of Section~\ref{sec:prism} uses equal joint and end-effector motion weights. Active frames satisfy $\Delta q_i\geq\max(Q_{0.05}(\Delta q),10^{-4})$. A stable ascending sort assigns evenly spaced ranks over active frames. The first frame copies the first available displacement. Gripper engagement is the episode-wise min--max normalization of the maximum gripper value across arms, with larger values denoting stronger engagement.

Acquisition uses a causal 11-frame average of positive engagement changes. Gripper activity uses a causal 13-frame average of gripper-displacement norms. Each signal is divided by its active-frame 92nd percentile plus $10^{-8}$ and clipped to $[0,1]$. For these normalized signals $a_i$ and $b_i$, the two trust factors are $\operatorname{clip}(1-0.70a_i,0.25,1)$ and $\operatorname{clip}(1-0.75b_i,0.25,1)$, respectively.

On active frames, the engagement-weighted term in Eq.~(\ref{eq:pps}) is supplemented by the disengaged baseline $0.35(1-g_i^{\mathrm{eng}})$. Inactive frames receive $0.5$. Scores are clipped to $[0,1]$ and smoothed with a centered seven-frame moving average using edge replication. An episode with no active frames retains the constant score $0.5$.

\begin{table}[!t]
\centering
\caption{Training configurations, real-robot execution settings, and offline PPS efficiency. PPS measurements use insertion trajectories and are independent of the policy backbone.}
\label{tab:implementation-settings}
\small
\setlength{\tabcolsep}{4pt}
\renewcommand{\arraystretch}{1.12}
\begin{tabular}{@{}p{0.32\linewidth}p{0.32\linewidth}p{0.32\linewidth}@{}}
\toprule
\rowcolor{simtableheader}Setting & StarVLA & $\pi_{0.5}$ \\
\midrule
VLM backbone & Qwen3-VL-4B-Instruct & PaliGemma (Gemma 2B) \\
Action expert & QwenPI\_v3, 36 layers, width 1,024 & Gemma 300M \\
Input image resolution & $224\times224$ & $224\times224$ \\
Robot state input & Yes & Yes \\
Action horizon $H$ & 15 & 15 \\
Action dimensions (dataset-specific) & 8 (single-arm RLBench) & 16 (MSSM); padded to 32 \\
\midrule
Training GPUs & $8\times$ NVIDIA H20 & $8\times$ NVIDIA H20 \\
Distributed training & DeepSpeed ZeRO-3 & Data parallel; FSDP axis 1 \\
Effective batch size & 192 & 192 \\
Batch per GPU & 12 & 24 \\
Gradient accumulation & 2 & 1 \\
Training updates & 15,000 & 15,000 \\
Optimizer & AdamW & AdamW \\
Adam $(\beta_1,\beta_2)$ & $(0.9,0.95)$ & $(0.9,0.95)$ \\
Optimizer $\epsilon$ & $10^{-8}$ & $10^{-8}$ \\
Weight decay & $10^{-8}$ & $10^{-10}$ \\
VLM learning rate & $10^{-5}$ & $5\times10^{-5}$ \\
Action expert learning rate & $10^{-4}$ & $5\times10^{-5}$ \\
Warmup updates & 500 & 500 \\
Learning-rate schedule & Cosine; minimum $5\times10^{-7}$ & Constant after warmup \\
Gradient norm clipping & 1.0 & 1.0 \\
\midrule
\multicolumn{3}{@{}l}{\textit{PRISM settings for $\pi_{0.5}$ connector insertion}} \\
CPR predictor & \multicolumn{2}{l}{Masked mean pooling; MLP width 256; SiLU; sigmoid output} \\
CPR loss coefficient & \multicolumn{2}{l}{$\lambda_{\mathrm{CPR}}=0.1$} \\
PSW parameters & \multicolumn{2}{l}{$\gamma=3$, $p=2$, $\tau_0=1$} \\
EMA decay & \multicolumn{2}{l}{0.999} \\
\midrule
\multicolumn{3}{@{}l}{\textit{Real-robot execution}} \\
Action execution frequency & \multicolumn{2}{l}{15\,Hz} \\
Predicted action chunk & \multicolumn{2}{l}{$H=15$ steps (1.0\,s)} \\
Executed actions per chunk & \multicolumn{2}{l}{$k=12$ steps before replanning (0.8\,s)} \\
\midrule
\multicolumn{3}{@{}l}{\textit{Offline PPS efficiency (three-run measurements)}} \\
PPS throughput & \multicolumn{2}{l}{$1.12\times10^{6}$ frames/s (computation only)} \\
State loading throughput & \multicolumn{2}{l}{$7.20\times10^{5}$ frames/s} \\
Temporal sorting throughput & \multicolumn{2}{l}{$1.67\times10^{7}$ frames/s} \\
\bottomrule
\end{tabular}
\end{table}

\section{Cross-Task PPS Computation and Visualization}
\label{app:pps-generality}
Figure~\ref{fig:pps-cross-task-1} illustrates PPS across nineteen tasks. Each row pairs a complete trace with ten observations selected near local score maxima in consecutive time intervals. Connectors indicate temporal correspondence. Scores use within-episode motion ranks, not absolute physical tolerances. Insertion uses replanning-step samples, whereas the remaining tasks use 30-Hz states with the smoothing settings above. RGB images provide context, not PPS inputs. All traces use the main text's color scale.

Beyond the three MSSM tasks, puzzle assembly involves picking up and aligning pieces with their matching locations. In screw placement, a small screw is picked up and deposited either inside a box or on a tray, forming two separate tasks. Water pouring transfers water between cups by lifting and tilting the source cup. Glue-bottle opening removes a threaded cap, while stamp capping aligns and replaces the cover on the stamp. Screwdriver-bit insertion seats a bit in an electric screwdriver, and USB insertion aligns a cable plug with a charger port.

The remaining tasks cover everyday object handling. A toothpick container is placed inside a tape ring, and a gripper model is pulled apart by the two arms. Gum-box opening lifts the package lid, while pen uncapping separates the cap from the pen. Cup placement lowers a cup onto a coaster, and box stacking places one red box on another. Carton closing folds the open flap of a white box shut. Scissors closing brings the open blades together. These demonstrations provide varied motion and engagement patterns for examining the cross-task behavior of PPS.

\begingroup
\setlength{\parskip}{0pt}
\captionsetup{font=small,skip=3pt}
\setlength{\baselineskip}{0pt}
\setlength{\lineskip}{1pt}
\noindent\includegraphics[width=\linewidth]{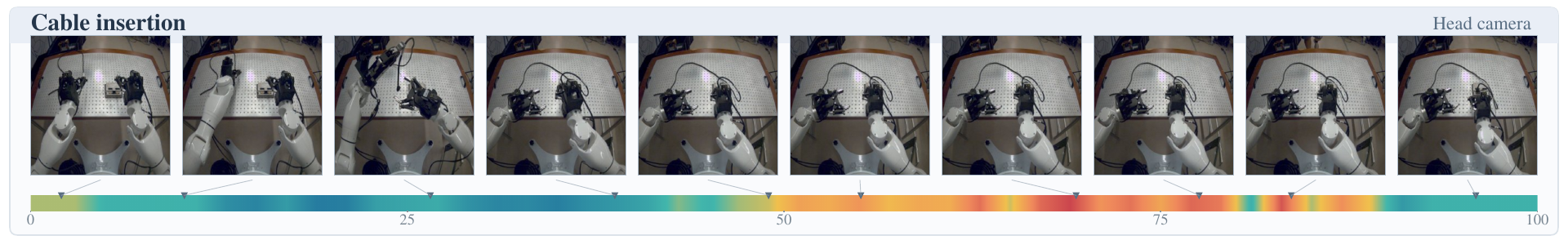}\par\penalty0
\noindent\includegraphics[width=\linewidth]{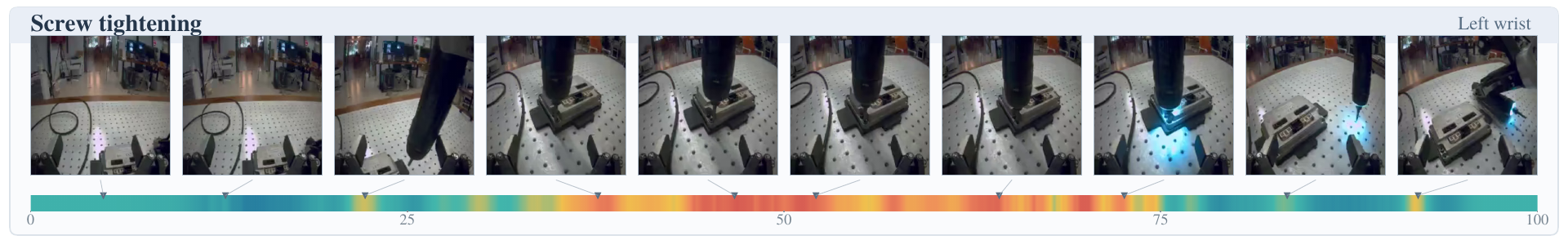}\par\penalty0
\noindent\includegraphics[width=\linewidth]{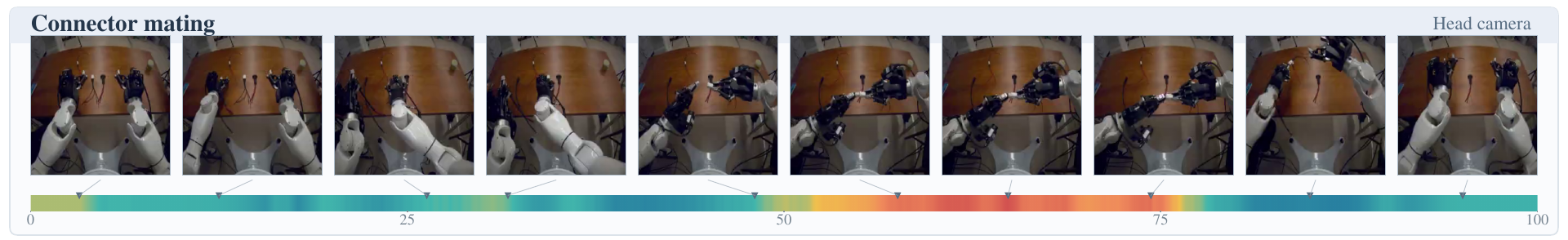}\par\penalty0
\noindent\includegraphics[width=\linewidth]{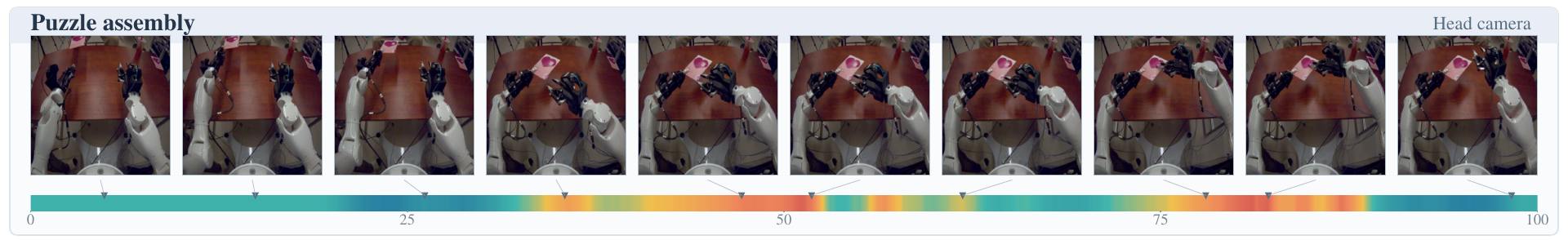}\par\penalty0
\noindent\includegraphics[width=\linewidth]{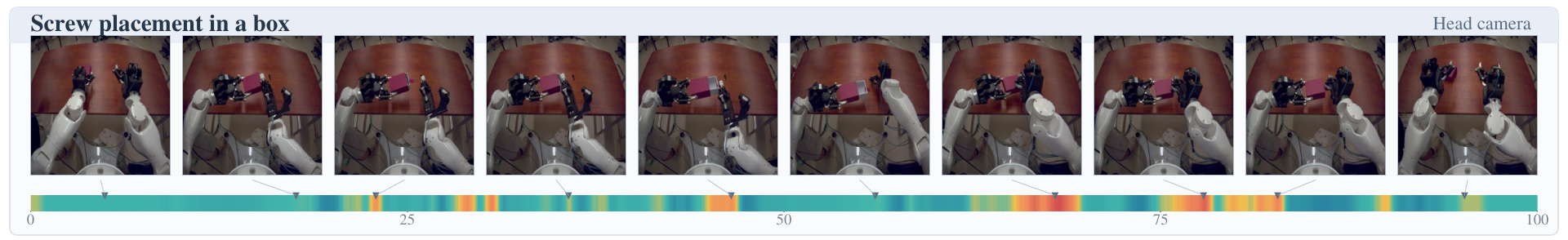}\par\penalty0
\noindent\includegraphics[width=\linewidth]{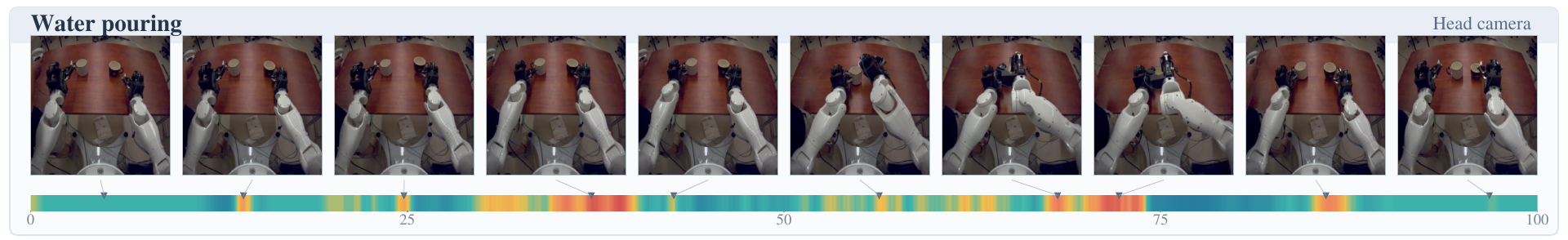}\par\penalty0
\noindent\includegraphics[width=\linewidth]{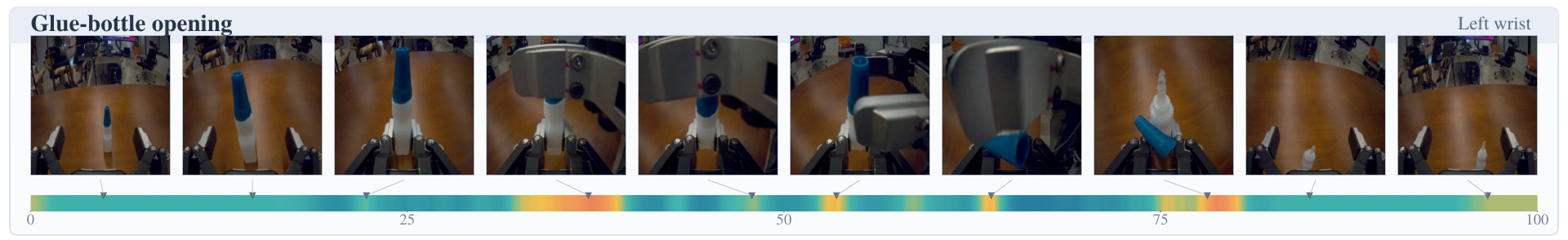}\par\penalty0
\noindent\includegraphics[width=\linewidth]{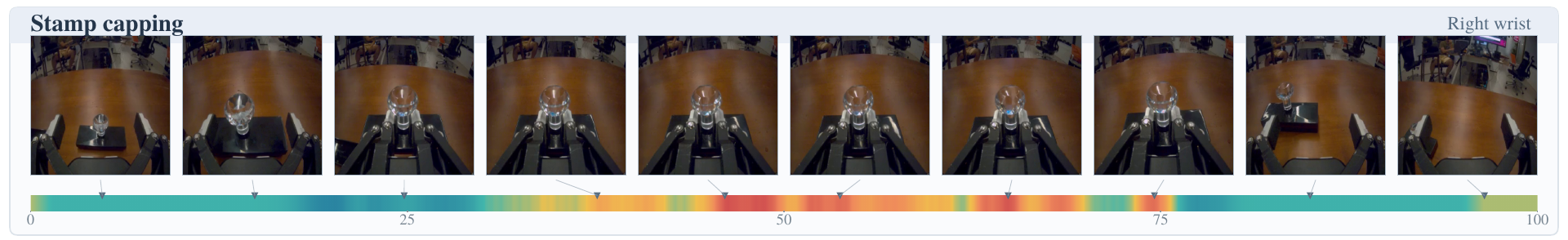}\par\penalty0
\noindent\includegraphics[width=\linewidth]{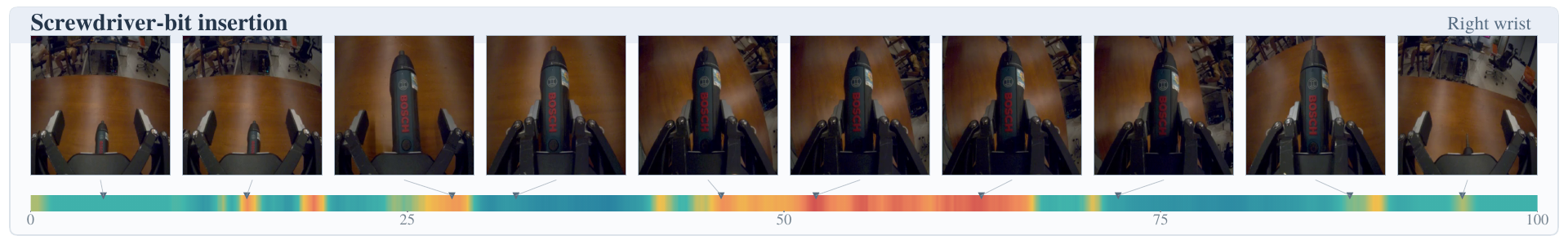}\par\penalty0
\noindent\includegraphics[width=\linewidth]{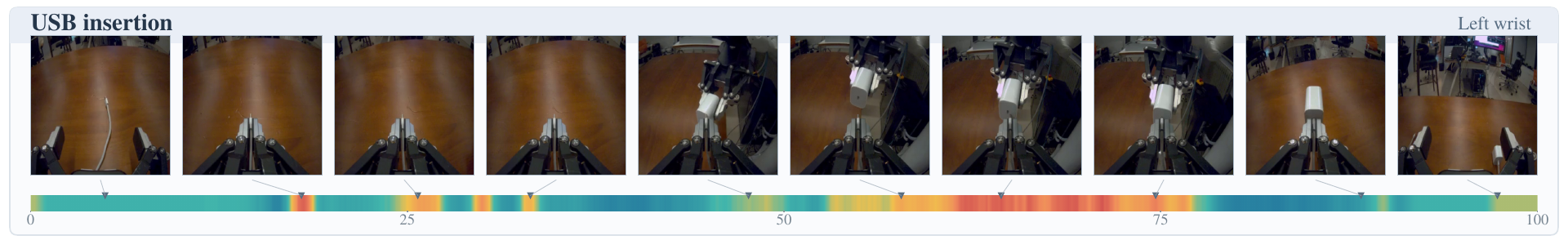}\par\penalty0
\noindent\includegraphics[width=\linewidth]{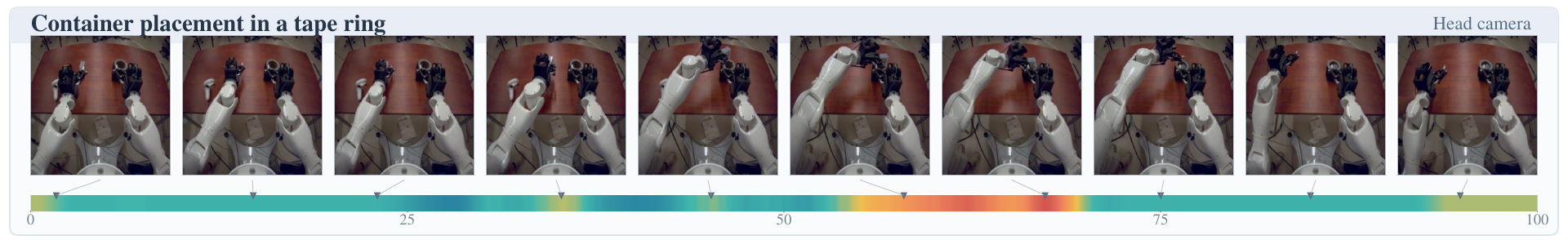}\par\penalty0
\noindent\includegraphics[width=\linewidth]{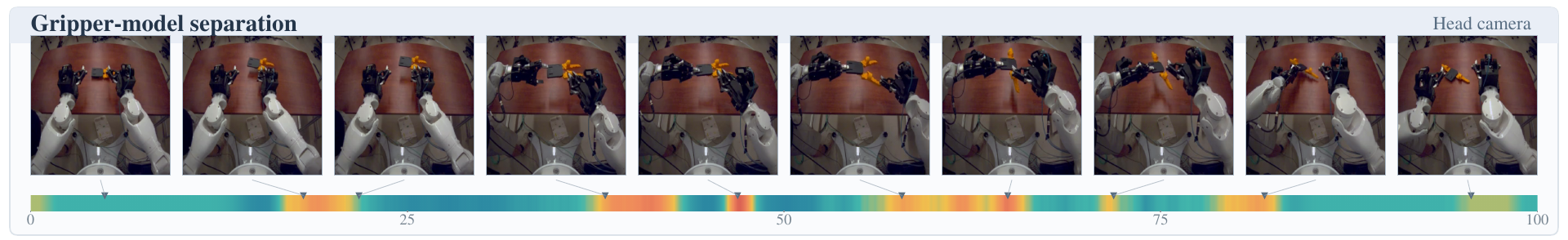}\par\penalty0
\noindent\includegraphics[width=\linewidth]{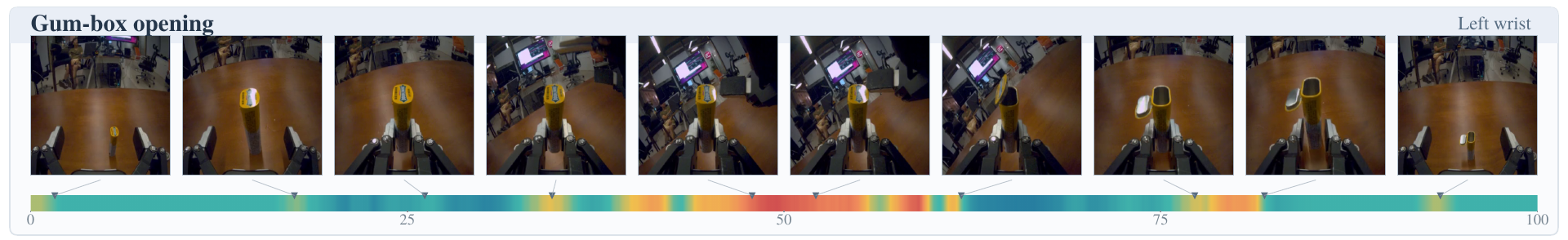}\par\penalty0
\noindent\includegraphics[width=\linewidth]{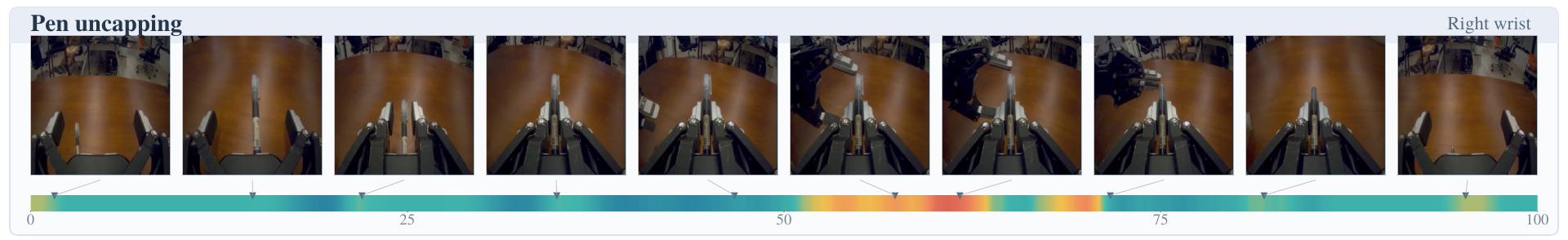}\par\penalty0
\noindent\includegraphics[width=\linewidth]{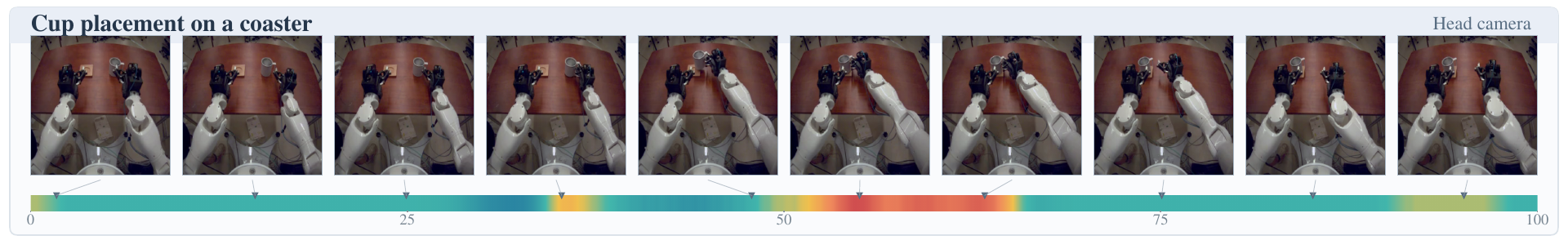}\par\penalty0
\noindent\includegraphics[width=\linewidth]{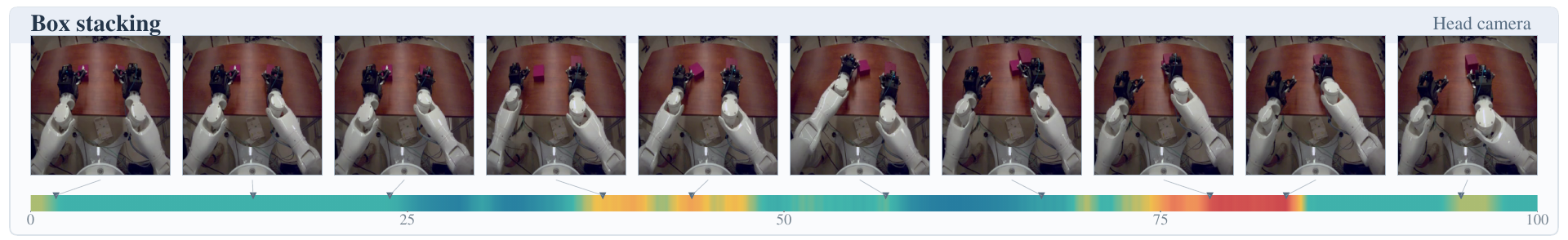}\par\penalty0
\noindent\includegraphics[width=\linewidth]{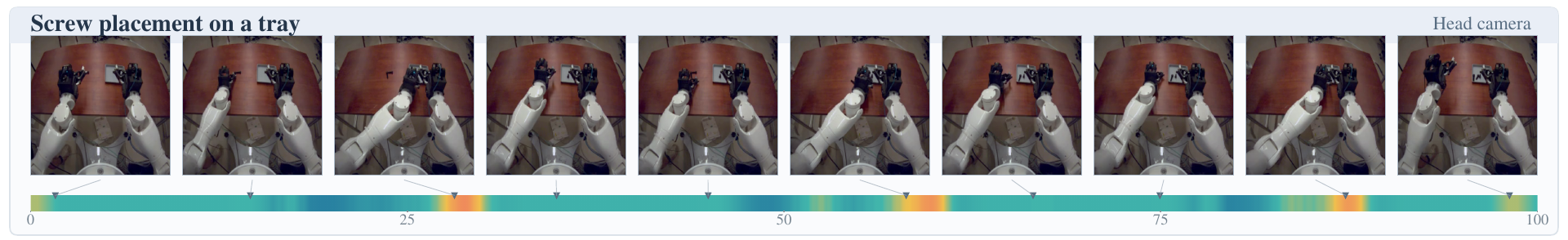}\par\penalty0
\noindent\includegraphics[width=\linewidth]{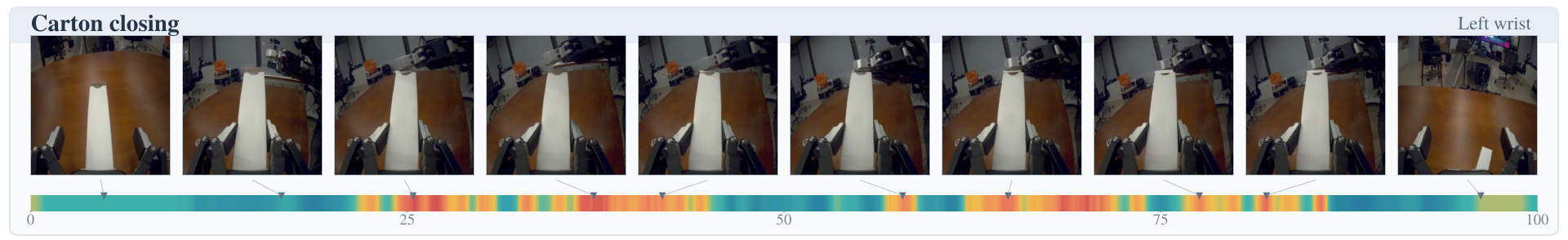}\par\penalty0
\endgroup
\noindent\begin{minipage}{\linewidth}
\includegraphics[width=\linewidth]{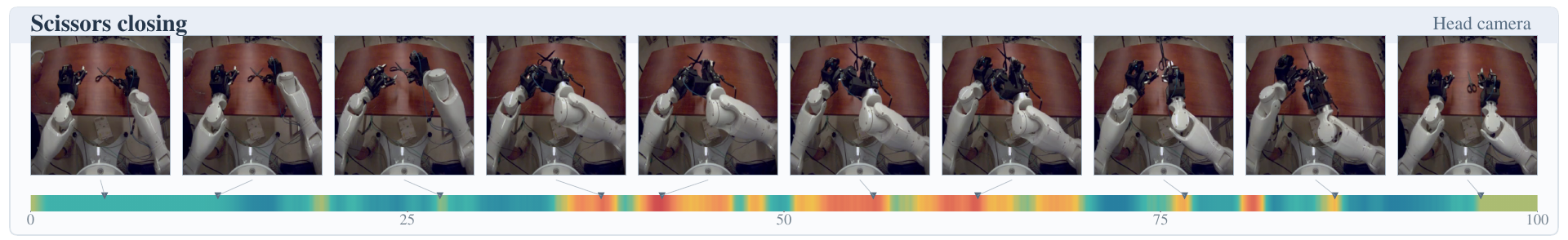}\par
\captionsetup{font=small,skip=3pt}
\captionof{figure}{\textbf{Cross-task PPS visualizations.} Nineteen task rows continue across pages. Each row shows ten observations and the full PPS trace, with horizontal coordinates denoting trajectory progress (\%).}
\label{fig:pps-cross-task-1}
\end{minipage}\par
\section{Task Protocols and Complete Results}
\label{app:mssm-protocols}
\paragraph{Robot platform, workspace, and objects.}
The Marvin dual-arm platform from TianJi has seven degrees of freedom per arm (14 in total) and standard two-finger parallel-jaw grippers. Visual observations come from a head camera and two wrist-mounted cameras, with camera hardware including the HIKROBOT MV-CS016-10GC. Demonstrations are recorded at 30\,Hz through VR teleoperation using a DPVR E4 Black headset and controllers with DPVR Assistant 4 software. Figure~\ref{fig:robot-hardware} shows the platform and teleoperation equipment. Real-robot execution settings are summarized in Table~\ref{tab:implementation-settings}.

\par\medskip\noindent\begin{minipage}{\linewidth}
\centering
\includegraphics[width=0.82\linewidth]{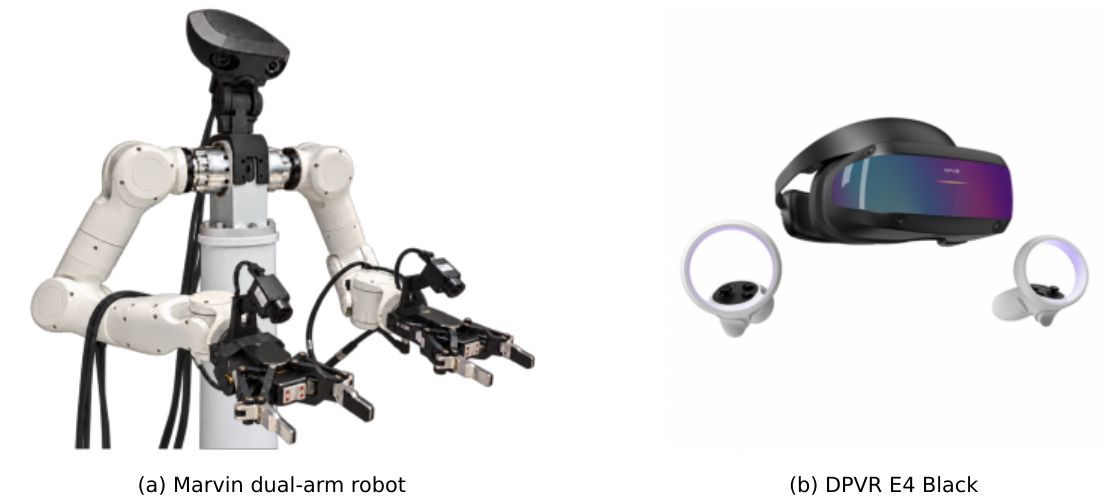}
\captionof{figure}{\textbf{Real-robot hardware.} Marvin dual-arm platform with mounted cameras (left) and DPVR E4 Black headset and controllers for teleoperation (right). The VR product image is from the \href{https://www.aiuto-jp.co.jp/products/product_5130.php}{DPVR distributor}.}
\label{fig:robot-hardware}
\end{minipage}\par\medskip

Objects are placed at predefined locations on a perforated metal workbench, and both arms start and finish each trial at standardized home configurations. Table~\ref{tab:mssm-objects} lists the task objects. The motor drive is placed on the right for connector insertion and near the center for screw tightening.
\begin{table}[htbp]
\centering
\caption{Objects used in the three MSSM tasks.}
\label{tab:mssm-objects}
\small
\setlength{\tabcolsep}{5pt}
\renewcommand{\arraystretch}{1.15}
\begin{tabular}{p{0.24\linewidth}p{0.68\linewidth}}
\toprule
\rowcolor{simtableheader}Task & Objects and specifications \\
\midrule
Drive connector insertion & Geek+ GS-SD-48V-20A-21D V01.01 dual-channel safety motor drive and a Hirose DF22 cable assembly with a matching drive receptacle. \\
Bimanual connector mating & Three connector pairs, used in order: white Hirose DF22 2-pin (0.2\,mm tolerance), white Hirose DF22 4-pin (0.1\,mm), and black Hirose DF60 (0.3\,mm). Male halves start on the left and female halves on the right. \\
Screw tightening & Bosch GO Gen 2 cordless screwdriver with a compatible bit; a loose screw pre-positioned in the lower-left threaded hole of the motor-drive housing. Measured alignment tolerance: 0.2\,mm. \\
\bottomrule
\end{tabular}
\end{table}

\paragraph{Tolerance calculation for Figure~\ref{fig:diagnosis}.}
The representative tolerance is the smallest single-sided margin among the specified constrained translation directions. A 94-mm gripper opening and 6.5-mm cable diameter give $(94-6.5)/2=43.75$\,mm for centered cable grasping. Handover permits offsets of $\pm10$, $\pm10$, and $\pm15$\,mm along the three translation directions, giving 10\,mm. The $22\times7$-mm socket and $21.6\times6.7$-mm plug give one-sided clearances of 0.20 and 0.15\,mm, so insertion uses 0.15\,mm under aligned orientation. For the other MSSM tasks, Table~\ref{tab:real-robot-main} reports the tightest measured tolerance: 0.1\,mm for bimanual mating, set by the white 4-pin pair, and 0.2\,mm for screw tightening.

\paragraph{Task stages.}
Table~\ref{tab:mssm-stage-descriptions} defines the evaluation stages in the same task order as Table~\ref{tab:real-robot-main}. During drive connector insertion, the left arm grasps the insulated cable behind the connector, and the right arm receives the housing across its narrow lateral faces. The left wrist camera observes the final plug--receptacle alignment. For bimanual mating, inward arm rotations and an approximately $90^\circ$ left-wrist adjustment align the keyed housings. Each connected pair is placed back on the workbench before proceeding to the next pair. Screw tightening uses the tool's push-to-start mechanism: the right arm maintains axial pressure after bit alignment, then withdraws and replaces the tool.
\begin{table}[htbp]
\centering
\caption{MSSM tasks, stages, and operation descriptions.}
\label{tab:mssm-stage-descriptions}
\small
\setlength{\tabcolsep}{5pt}
\renewcommand{\arraystretch}{1.20}
\begin{tabular}{p{0.21\linewidth}p{0.23\linewidth}p{0.48\linewidth}}
\toprule
\rowcolor{simtableheader}Task & Stage & Description \\
\midrule
Drive connector insertion & S1: Cable pickup & The left arm picks up the cable behind its connector housing. \\
& S2: Bimanual handover & The left arm transfers the cable to the right arm. \\
& S3: Connector insertion & The right arm aligns the cable connector with the receptacle and inserts it. \\
\midrule
Bimanual connector mating & S1: Left-arm pickup & The left arm grasps the male half of the first connector pair. \\
& S2: Right-arm pickup & The right arm grasps the female half of the first connector pair. \\
& S3: First pair & Align and mate the Hirose DF22 2-pin pair, then place it back on the workbench. \\
& S4: Second pair & Pick up, align, and mate the Hirose DF22 4-pin pair, then place it back. \\
& S5: Third pair & Pick up, align, and mate the Hirose DF60 pair, then place it back. \\
\midrule
Screw tightening & S1: Tool pickup & The right arm grasps the cordless screwdriver around its body. \\
& S2: Assisted reorientation & The left gripper nudges and reorients the held tool until its bit points approximately downward. \\
& S3: Tightening and return & Align the bit with the pre-positioned screw, press downward to tighten it, then withdraw and return the screwdriver to its original location. \\
\bottomrule
\end{tabular}
\end{table}
\section{Additional Ablations and Visualizations}
\begin{figure}[!ht]
 \centering
 \includegraphics[width=0.76\linewidth]{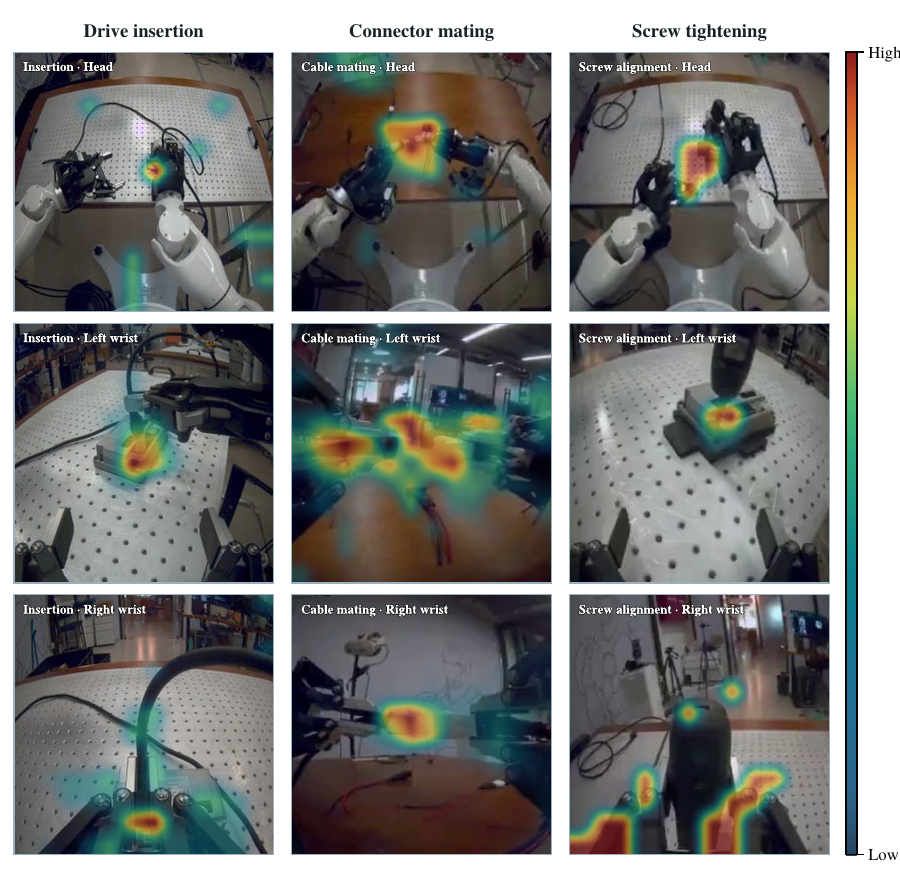}
 \caption{\textbf{CPR sensitivity across three tasks and three camera views.} Each column shows one task. Rows show the head, left-wrist, and right-wrist cameras at the same selected observation within that task. Labels identify the action and camera. Colors indicate relative within-view sensitivity, not a shared absolute response scale. Selected views and heatmaps are jointly rotated for readability.}
 \label{fig:cpr-multiview}
\end{figure}

\paragraph{Multi-view CPR sensitivity.}
Figure~\ref{fig:cpr-multiview} extends Figure~\ref{fig:mechanism_discussion}(d) to the head and both wrist cameras for each MSSM task. We blur one image patch at a time while holding the other views, state, and instruction fixed, and measure the absolute change in predicted precision demand. The $12\times12$ overlapping sampling grid uses patches approximately one sixth of the image width and height and a Gaussian blur radius of 18 pixels. Maps use the same color palette and response-dependent opacity as Figure~\ref{fig:mechanism_discussion}(d), normalized by each map's maximum. Wrist views show localized responses near the drive connector interface, connector mating interface, and screwdriver contact region, while some head and opposite-wrist views also respond to surrounding scene content. The comparison illustrates view-dependent sensitivity rather than uniform localization across cameras.
\begin{figure}[!ht]
  \centering
  \begin{minipage}[c]{0.47\linewidth}
  \small
  \textbf{Amplification strength.}
  Figure~\ref{fig:gamma-sweep} reports success rates of $0.80$, $0.85$, $0.95$, $0.90$, and $0.85$ for $\gamma=1,\ldots,5$ on drive connector insertion. We choose $\gamma=3$, which performs best in this sweep. Qualitatively, increasing $\gamma$ from 1 to 3 yields more effective recovery from unfavorable pre-alignment configurations. Larger values reduce success, suggesting that excessive emphasis on high-demand samples can make supervision less balanced.
  \end{minipage}\hfill
  \begin{minipage}[c]{0.50\linewidth}
  \includegraphics[width=\linewidth]{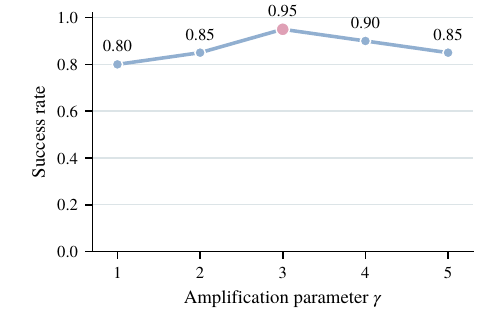}
  \end{minipage}
  \caption{PSW amplification on drive connector insertion. Success peaks at $\gamma=3$.}
  \label{fig:gamma-sweep}
\end{figure}

\subsection{Flow-Time Perturbation Diagnostics}
\label{sec:flow-diagnostics}
\paragraph{Equal-energy velocity perturbations.}
We evaluate the baseline $\pi_{0.5}$ on ten held-out drive-insertion episodes, using two observations from each of the pickup, handover, and insertion stages. Observations are selected at fixed within-stage fractions using gripper transitions, independently of PPS. For each of the 60 observations, we use two initial-noise seeds, two random perturbation directions, and paired positive and negative perturbations. We retain the native ten-step Euler sampler and perturb its velocity in one of five equal-width flow-time intervals. The perturbation is constant within each interval and acts on the 16 active action dimensions over the 15-step action chunk. Its integrated mean squared magnitude is either $0.0004$ or $0.0025$, identical across intervals within each energy setting.

\begin{figure}[!ht]
 \centering
 \includegraphics[width=\linewidth]{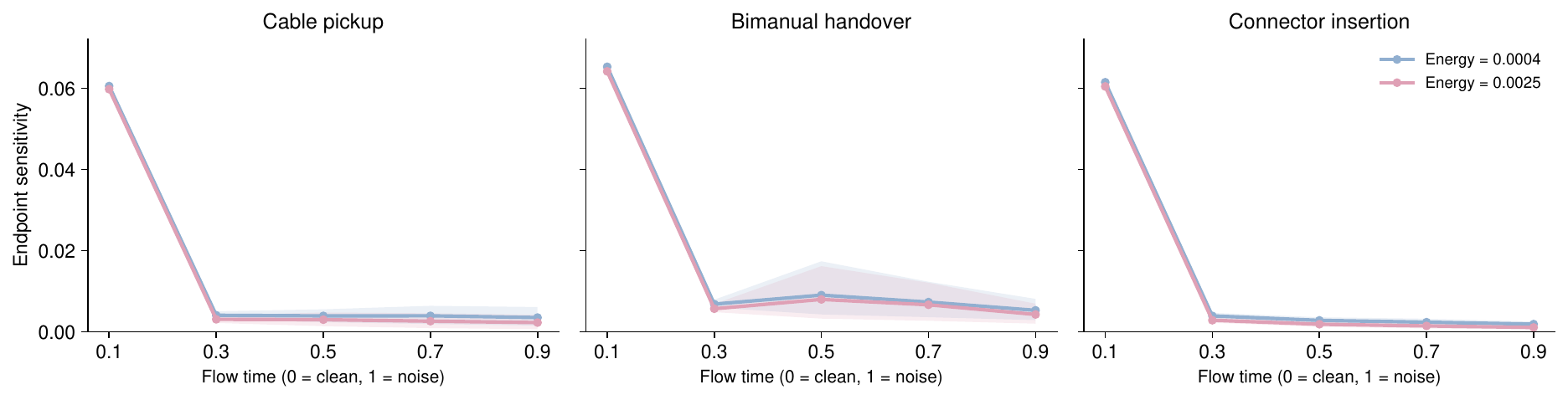}
 \caption{\textbf{Equal-energy perturbations across flow time.} Curves show unweighted endpoint action MSE per unit perturbation energy on ten held-out episodes. Shaded bands are 95\% episode-bootstrap intervals. The clean-end interval produces the largest response in each operation stage at both energy settings.}
 \label{fig:equal-energy-flow}
\end{figure}

The endpoint response is the mean squared difference between perturbed and unperturbed generated action chunks, in normalized action coordinates, divided by perturbation energy. Figure~\ref{fig:equal-energy-flow} reports episode-averaged responses and 95\% episode-bootstrap intervals. Both energy settings yield the largest response in the clean-end interval, $\tau\in[0,0.2]$, across all three stages. Their clean-end response magnitudes are similar, indicating a shared flow-time propagation effect. For Figure~\ref{fig:diagnosis}(b), we multiply the response at energy $0.0025$ by the dimensionless inverse-tolerance ratios $1$, $4.375$, and $291.67$. This constructs a tolerance-weighted proxy from the measured action-space response and the stage tolerances, rather than measuring Cartesian displacement directly.

\section*{AI Use Statement}
We used AI tools for writing and language polishing. The authors reviewed all AI-assisted outputs and take responsibility for the final content.

\end{document}